\documentclass{article}
\usepackage{iclr2027_arxiv,times}
\usepackage[T1]{fontenc}
\usepackage[utf8]{inputenc}

\usepackage{amsmath,amsfonts,bm}

\def\eqref#1{equation~\ref{#1}}

\def\1{\bm{1}}

\DeclareMathAlphabet{\mathsfit}{\encodingdefault}{\sfdefault}{m}{sl}
\SetMathAlphabet{\mathsfit}{bold}{\encodingdefault}{\sfdefault}{bx}{n}

\usepackage{amsmath,amssymb,array,booktabs,float,graphicx,microtype,longtable}
\usepackage[export]{adjustbox}
\renewcommand{\eqref}[1]{\textup{(\ref{#1})}}
\usepackage{hyperref}
\hypersetup{hidelinks}
\usepackage{needspace}  %
\usepackage{placeins}  %
\usepackage{xurl}  %
\newcommand{\CodeURL}{{\def\UrlBreaks{\do\/\do\-}\def\UrlBigBreaks{}\url{https://github.com/Soren-Shangzhen/Approximating-Softmax-in-Pretrained-LLMs-Model-Sensitivity-and-Kernel-Acceleration}}}
\graphicspath{{figures/}}
\title{Approximating Softmax in Pretrained LLMs:\\Model Sensitivity and Kernel Acceleration}
\author{Shangzhen Zhu$^{1}$, Muyan Hu$^{1}$ \& Tomasz Kozlowski$^{1,2}$ \\
$^{1}$University of Illinois Urbana-Champaign \\
$^{2}$National Centre for Nuclear Research, Otwock, Poland \\
\texttt{\{szhu48,muyanhu2,txk\}@illinois.edu}}
\iclrfinalcopy

\newcommand{\AppendixFigure}[4]{%
  \begin{figure}[!htbp]%
    \centering
    \includegraphics[max width=\textwidth,max height=0.88\textheight]{#1}%
    \caption{\textbf{#2} #3}
    \label{#4}
  \end{figure}
}

\begin{document}
\maketitle
\begingroup\makeatletter\renewcommand{\thefootnote}{}\long\def\@makefntext#1{\noindent #1}\makeatother
\footnotetext{\raggedright Code and reference results: \CodeURL}\endgroup
\begin{abstract}
On NVIDIA Blackwell B200, tensor-core throughput outpaces special-function
exponential throughput by more than two orders of magnitude, exposing exponential
evaluation in fused attention kernels.  A pretrained
Transformer, however, may not need it evaluated accurately at every element.

We characterize what a pretrained model does need by approximating softmax at inference in ten
frozen decoder-only models (0.5B--72B).  The number of positions the softmax map assigns probability to and
within-row resolution can be cut substantially, yet uniform weighting of the same
positions is damaging.  Where a fixed resolution budget is placed matters as much
as its size, with resolution near the row maximum consistently favored.
Perturbations matched on scalar distortion produce model-dependent responses of
opposite sign.

These findings motivate \mbox{Rowmax-PoT}, a coarse logarithmic weight representation
anchored at each row maximum, and \mbox{Rowmax-H15}, its hardware specialization in
FlashAttention-4.  On B200, the patched FP8 attention forward is 12.4\% faster at causal 8K
and 25.8\% faster at non-causal 8K in host-side call-latency measurements; board energy per
forward falls by 8.4\% at causal 16K.  Measured separately on the BF16 kernel path at 2K,
\mbox{Rowmax-H15} increases perplexity by 0.091--0.492\% across five models from three families.
\end{abstract}

\section{Introduction}
\label{sec:m-intro}

Softmax attention assigns an exponentially weighted probability to every unmasked position
of a row, at a precision that may exceed what a pretrained model needs.  We study which
properties of this computation can be relaxed at inference time in frozen decoder-only
models, and whether the resulting approximations can accelerate a fused attention kernel
without retraining.  Each experiment changes one property and fixes the others
(Table~\ref{tab:dof}).

Interventions on ten frozen models (0.5B--72B) give three findings.  First, support size and grid
resolution can be reduced substantially in the tested settings.  In Qwen2.5-1.5B, retaining
the largest half of each attention row increases perplexity by \(0.131\%\).  A separate
single-layer control uses identical mean-threshold supports in its two conditions: uniform
weights instead of softmax weights increase NLL at every tested target layer.  Across ten
models, a full-range grid of 32 intervals, finer near the row maximum (\(K=32,R=4\);
Section~\ref{sec:m-defs}), stays below a 1\% perplexity increase in every point estimate
(Section~\ref{sec:m-amount}).  Second, at a fixed number of intervals, finer intervals near the row maximum lower NLL in all ten point
estimates (nine of ten 95\% intervals exclude zero), although in Qwen2.5-1.5B the
unweighted error of the log weights increases (Section~\ref{sec:m-allocation}).  Third,
attention JSD tracks damage but does not determine it: for perturbations calibrated to the
same mean row-wise JSD, flattening is more damaging than sharpening in seven models and less
damaging in two
(Section~\ref{sec:m-structure}).

Exponential evaluation can limit fused attention throughput on current GPUs: Blackwell tensor cores
deliver 8192 matrix operations per clock per SM against 16 exponentials from the
special-function unit, and FlashAttention-4 (FA4) already uses polynomial approximations
for some \texttt{exp2} evaluations \citep{flashattention4-2026}.  Our observations
motivate a coarse approximation.  It keeps approximately exponential weights, because
uniform weights were damaging, and it places resolution relative to the running row maximum
that the online-softmax loop tracks, as the allocation results suggest; Section~\ref{sec:m-anchor}
compares anchors directly.  Because equal JSD gave different losses, we evaluate the kernel
by model NLL in addition to pointwise error.  Rowmax-PoT rounds the base-two logarithm of
each weight, measured from the row maximum, to an integer; Rowmax-H15, its FA4
specialization, replaces \texttt{exp2} by values \(\{1,1.5\}\times2^k\) computed with rounding
and integer bit operations (Section~\ref{sec:m-realization}).

\paragraph{Contributions.}
\begin{enumerate}
  \item \textbf{Properties that can and cannot be relaxed.}  In frozen models, support size and
  the number of grid intervals can be reduced substantially in the tested settings, while
  uniform weights on the same support increase NLL at every tested layer
  (Section~\ref{sec:m-amount}).
  \item \textbf{Allocation and model-dependent sensitivity.}  The placement of a fixed
  number of grid intervals changes NLL, and perturbations calibrated to the same attention
  JSD produce model-dependent losses (Sections~\ref{sec:m-allocation}--\ref{sec:m-geometry}).
  \item \textbf{Rowmax-H15 inside FA4.}  The BF16 kernel raises perplexity by 0.091--0.492\% in
  five models; under different conditions, the FP8 attention forward is 12.4\% faster at
  causal 8K and uses 8.4\% less board energy per forward at causal 16K
  (Section~\ref{sec:m-realization}).
\end{enumerate}

\section{Related Work}
\label{sec:m-related}

\textbf{Quantized attention probabilities and the anchor.}  Post-training quantization of
vision transformers places the non-uniform post-softmax map on twin uniform grids
\citep{ptq4vit-2022}, a \(\log_2\) grid \citep{fq-vit-2022} or a \(\log_{\sqrt2}\) grid
\citep{repq-vit-2023}.  PTQ4SAM searches the power-of-two base offline \citep{ptq4sam-2024},
and APQ-ViT calibrates so that the power-law tail survives \citep{apq-vit-2022}; both
indicate that the large entries must be resolved.  These grids are anchored to a
row-independent scale, fixed at \(p=1\) or calibrated per layer, so their phase relative to
each row maximum varies across rows; RepQ-ViT halves the spacing, which spends resolution.
EXAQ already anchors at the row maximum: it uniformly quantizes the \emph{input} to the
exponential after subtracting the row maximum, with a clipping value calibrated offline
\citep{exaq-2024}.  Section~\ref{sec:m-anchor} compares the two anchors.

\textbf{Approximate exponentials and model-level validation.}  I-BERT approximates the
exponential with an integer polynomial and I-ViT with a base-2 shift and a linear segment;
both recover accuracy by quantization-aware fine-tuning \citep{i-bert-2021,i-vit-2023}.
Softermax and ITA design base-2 or integer softmax together with custom hardware
\citep{softermax-2021,ita-2023}, ConSmax replaces the row maximum and the denominator with
learned normalization parameters \citep{consmax-2024}, and IntAttention uses a clipped
32-entry lookup table without training on edge CPUs \citep{zhong2026intattention}.  In the FA4
implementation, we retain the online row-maximum update and normalization and change only the
exponential evaluation.  Model parameters remain frozen.  I-BERT and FA4 justify their approximations by an error below
storage precision; the per-element error of Rowmax-H15 is far larger than one BF16 unit in
the last place, so we validate it by model loss.  \citet{schraudolph1999fast} approximates
exponentiation by an affine map into the IEEE-754 exponent and mantissa bits, interpolating
linearly between powers of two; Rowmax-H15 is a coarse specialization of this construction
(Appendix~\ref{app:schraudolph}).

\textbf{Fused kernels and operand quantization.}  FlashAttention and its successors provide
the fused online-softmax kernel whose running row maximum our lattice uses
\citep{flashattention-2022,flashattention2-2023,flashattention3-2024}.  SageAttention 1--3
and INT-FlashAttention quantize the matrix-multiply operands inside fused kernels, and
MXAttention quantizes the unnormalized exponentials after exact evaluation
\citep{sageattention-2024,sageattention2-2024,sageattention3-2025,int-flashattention-2024,mxattention-2026}.
These methods change the operand format and keep an exact exponential; Rowmax-H15 changes
the exponential between the two matrix multiplies, and the two could in principle be
combined.  FA4 \citep{flashattention4-2026} is the kernel we modify and our controlled kernel
baseline.  Two Meta engineering reports address the same special-function unit:
Low-Precision FA4 extends FA4 with MXFP8 operands on GB300, keeps the FP32 softmax and
reports that the SFU-bound softmax becomes exposed behind faster matrix multiplies
\citep{lp-fa4-mxfp8-2026}, and GDPA replaces softmax with element-wise activations evaluated
by an ALU polynomial \citep{gdpa-2026}.

\textbf{Concurrent work.}  EFQ-Softmax \citep{efq-softmax-2026} also removes the per-element
exponential inside a FlashAttention-style loop, and the positive part of its E2M1 code set,
\(\{1,1.5\}\times2^k\), coincides with our lattice.  The designs differ in anchor
(microscaling-block maximum versus attention-row maximum), rounding (affine thresholds
versus round-to-nearest-even), parameters (an offline two-parameter search versus none) and
measured scope (a kernel's vector stage versus the whole attention-forward kernel).
Table~\ref{tab:rw-compare} compares the closest methods; our literature search is not
exhaustive.

\section{Framework and Experimental Setup}
\label{sec:m-framework}

\begin{table}[t]
\centering
\small
\setlength{\tabcolsep}{3pt}
\caption{\textbf{Intervention dimensions.}  Each experiment varies the listed property and
fixes the others.}
\label{tab:dof}
\begin{tabular}{@{}llll@{}}
\toprule
Degree of freedom & Intervention (others fixed) & Endpoint & Where \\
\midrule
Support & top-\(k\), mean threshold & \(\Delta\mathrm{NLL}\) & \S\ref{sec:m-amount}, Fig.~\ref{fig:m-amount}A \\
Within-support weighting & uniform vs.\ softmax, same support & \(D_W(\ell)\) & \S\ref{sec:m-amount}, Fig.~\ref{fig:m-amount}B \\
Intervals \(K\); weight map & full-range grid; exp.\ vs.\ linear map & \(\Delta\mathrm{NLL}\) & \S\ref{sec:m-amount}, Fig.~\ref{fig:m-amount}C, Fig.~\ref{fig:amount-geometry-supp} \\
Allocation \(R\) & \(K=21\), \(R\in[0.25,8]\) & \(D_R(21)\), \(I_R\) & \S\ref{sec:m-allocation}, Fig.~\ref{fig:m-allocation} \\
Reconstruction & upper edge, nearest, interpolation & \(\Delta\mathrm{NLL}\) & \S\ref{sec:m-recon} \\
Exponential evaluation & Rowmax-PoT; Rowmax-H15 in FA4 & \(\Delta\mathrm{NLL}\), latency & \S\ref{sec:m-recon}, \S\ref{sec:m-realization} \\
\bottomrule
\end{tabular}
\end{table}

\subsection{Models and evaluation protocol}
\label{sec:m-setting}

All pretrained parameters are frozen.  We replace only the map
from a row of attention scores to its normalized weights, in every head and layer
unless stated otherwise.  For a query row with valid key set \(\mathcal V\) and
scores \(s_j\), the \emph{softmax} is
\begin{equation}
  \Delta_j=\max_{k\in\mathcal V}s_k-s_j,
  \qquad
  p_j=\frac{\exp(-\Delta_j)}{\sum_{k\in\mathcal V}\exp(-\Delta_k)},
  \label{eq:m-exact}
\end{equation}
recomputed from the scores produced under the current intervention.  We use attention
support to mean the set of valid key positions assigned nonzero attention weight; full support
means that every position permitted by the original attention mask retains nonzero weight.  The primary
endpoint is WikiText-103 next-token NLL: the test stream is cut into 97 aligned
blocks of 2048 tokens (198,559 predictions), \(\Delta\mathrm{NLL}\) is measured
against the softmax baseline within each model, and perplexity change is
\(\exp(\Delta\mathrm{NLL})-1\).  Contrasts are paired over blocks with a 5,000-replicate
percentile bootstrap (seed 0); a contrast is \emph{resolved} when its 95\% interval
excludes zero.  The suite has ten decoder-only models from four families: a core set
of eight (Qwen2.5-0.5B/1.5B/3B, Llama-3.2-1B/3B, Llama-3.1-8B, Gemma-2-2B,
Mistral-7B-v0.3) and two larger models (Qwen2.5-72B, Llama-3.1-70B) that run a subset of
the experiments; Qwen2.5-1.5B receives the detailed diagnostic sweeps.  Characterization
(Section~\ref{sec:m-char}) runs in PyTorch 2.5.1 with BF16 weights and FP32 score-to-probability
computation, on one GPU (core models) or three A100-SXM4-80GB GPUs (the two larger models).  Models are
analyzed individually, and no pooled cross-model effect is estimated.  The primary
statistical contrasts are \(D_W\), \(D_R(21)\), \(D_T(J_0)\), \(D_T(10J_0)\) and
\(G_{\mathrm{gap}}(10J_0)\) (Section~\ref{sec:m-defs}); the others are supporting or secondary
(Appendix~\ref{sec:methods-statistics}).

\subsection{Key definitions}
\label{sec:m-defs}

\paragraph{Full-range grid.}  With shifted scores \(u_j=s_j-\min_k s_k\in[0,C]\) and
\(C=\max s-\min s\), each row's range \([0,C]\) is divided into \(K\) intervals (\(K+1\)
boundaries)
\begin{equation}
  e_a=C\,\frac{1-\rho^a}{1-\rho^K},\qquad \rho=R^{-1/(K-1)},\qquad a=0,\ldots,K,
  \label{eq:m-grid}
\end{equation}
so that the width of the lowest-score interval is \(R\) times that of the interval at the
row maximum (\(R=1\): uniform widths); \(R>1\) places finer intervals near the row maximum.
Weights are exponential, with nearest-boundary reconstruction unless stated otherwise
(Appendix~\ref{sec:methods-quantization}).

\paragraph{Same-support weighting.}  For each target layer \(\ell\) of Qwen2.5-1.5B, both
conditions use softmax attention below \(\ell\), so they reach \(\ell\) with the same scores
and the same mean-threshold support (scores above the mean valid score).  At \(\ell\) they differ only between softmax and uniform
weights on that support; later layers return to softmax attention.  \(D_W(\ell)\) is the
paired uniform-minus-softmax \(\Delta\mathrm{NLL}\) (Appendix~\ref{sec:methods-support}, Eq.~\eqref{eq:dw-layer}).

\paragraph{Allocation contrast.}  At a fixed number of intervals \(K\),
\begin{equation}
  D_R(K)=\mathbb E_b\!\left[\Delta\mathrm{NLL}_{b,K,R=1}-\Delta\mathrm{NLL}_{b,K,R=4}\right],
  \qquad I_R=D_R(16)-D_R(32),
  \label{eq:m-dr}
\end{equation}
so \(D_R>0\) means that finer intervals near the row maximum lower NLL.

\paragraph{Matched distortion.}  \(J_0\) is the mean row-wise JSD that an octave power-of-two
perturbation of Eq.~\eqref{eq:m-exact} produces in a given model
(Appendix~\ref{sec:methods-pot}).  Flattening (\(A^-\)) and sharpening (\(A^+\)) apply softmax
to scores scaled by \(\alpha<1\) and \(\alpha>1\), with \(\alpha\) set to reach mean JSD \(J_0\)
or \(10J_0\), and
\begin{equation}
  D_T(J)=\mathbb E_b\!\left[\Delta\mathrm{NLL}_{b,A^-(J)}-\Delta\mathrm{NLL}_{b,A^+(J)}\right].
  \label{eq:m-dt}
\end{equation}
so \(D_T>0\) means that flattening raises NLL more than sharpening.  Each \(\alpha\) is found by bisection on the divergence alone, without access to NLL, on
eight calibration blocks and then frozen (Appendix~\ref{sec:methods-matching}); equal JSD is
not re-verified on the 97 evaluation blocks.  Matching equalizes the aggregate JSD, not its
distribution over layers, heads and rows, so a nonzero \(D_T\) shows that this scalar summary
is insufficient.  Mass-partition restoration (``rescue'')
rescales the set carrying 90\% of the softmax mass and its complement back to their softmax mass; \(G_{\mathrm{gap}}=D_{\mathrm{orig}}-D_{\mathrm{rescue}}\)
is the signed change in the contrast, and a positive value does not imply a smaller \(|D|\)
(Appendix~\ref{sec:methods-rescue}).

\section{Sensitivity of Frozen Models to Softmax Approximations}
\label{sec:m-char}

\subsection{Support, resolution and relative weights}
\label{sec:m-amount}

\begin{figure}[t]
  \centering
  \includegraphics[width=\textwidth]{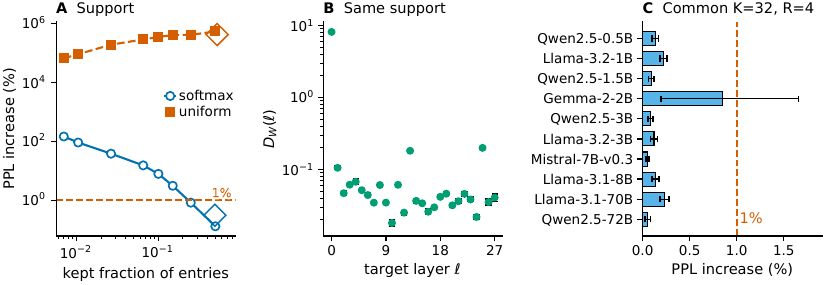}
  \caption{\textbf{Attention support, weighting, and a 32-interval grid.}  (A)~Support sweep in
  Qwen2.5-1.5B with softmax (blue) or uniform (orange) weights; both axes are logarithmic.
  Diamonds mark the rule that retains scores above the row mean, plotted at each run's own kept
  fraction (0.493 for softmax; 0.526 for uniform).  The interventions run in every layer and
  change later scores, so panel A does not hold support fixed between weighting rules.
  (B)~Separate single-layer control in Qwen2.5-1.5B with identical mean-threshold supports; the
  y-axis is logarithmic.  \(D_W(\ell)\) is NLL with uniform weights minus NLL with softmax weights
  at target layer \(\ell\), so positive values mean uniform weighting increases NLL.  All 28
  pointwise 95\% paired-bootstrap intervals exclude zero.  (C)~Perplexity increase across ten
  models using 32 full-range intervals, narrower near the row maximum (\(R=4\)).  Whiskers are
  95\% paired-bootstrap intervals; the dashed line marks a 1\% increase.}
  \label{fig:m-amount}
\end{figure}

Support size and grid resolution can be reduced substantially in the tested settings, up to a
failure regime at small support sizes.  In Qwen2.5-1.5B, keeping the largest half of each row with
its softmax weights raises perplexity by \(0.131\%\), keeping a quarter raises it by
\(0.828\%\), and a nominal retained fraction of 1\% raises it by \(90.478\%\) (Figure~\ref{fig:m-amount}A).
With full support, increasing \(K\) reduces \(\Delta\mathrm{NLL}\) for the \(R=1\) grid with
upper-edge reconstruction, reaching \(0.00981\) at \(K=32\) (Figure~\ref{fig:amount-geometry-supp}).  With
nearest-boundary reconstruction, the common \(K=32,R=4\) grid stays below a 1\%
perplexity increase in all ten point estimates; only the interval of Gemma-2-2B spans 1\%
(Figure~\ref{fig:m-amount}C; Table~\ref{tab:crossmodel-summary}).

In the single-layer same-support control on Qwen2.5-1.5B, uniform weighting increases NLL at
all 28 target layers: every pointwise 95\% interval for \(D_W(\ell)\) excludes zero, without
adjustment for multiplicity (Figure~\ref{fig:m-amount}B; Appendix~\ref{sec:results-amount}).  The tested linear
level-to-weight map also degrades the model severely: the paired linear-minus-exponential
contrast is \(8.33\) \([8.15,8.51]\) at \(K=32\), a result about this map only
(Figure~\ref{fig:amount-geometry-supp}).

\subsection{Allocation of a fixed number of intervals}
\label{sec:m-allocation}

\begin{figure}[t]
  \centering
  \includegraphics[width=\textwidth]{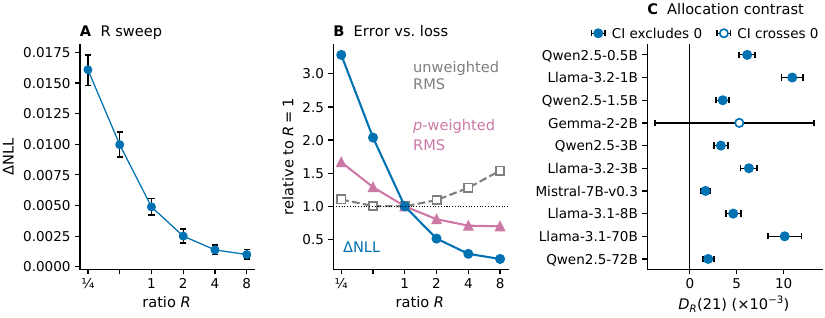}
  \caption{\textbf{Allocation of a fixed number of intervals.}  \(R>1\) makes intervals narrower
  near the row maximum; \(R=1\) gives uniform widths.  (A)~\(\Delta\mathrm{NLL}\) in Qwen2.5-1.5B
  as \(R\) varies at \(K=21\); whiskers are 95\% paired-bootstrap intervals.
  (B)~\(\Delta\mathrm{NLL}\) and RMS log-weight errors relative to their values at \(R=1\).  The
  errors use scores from the softmax baseline; \(\Delta\mathrm{NLL}\) evaluates the approximation
  in every layer.  From \(R=1\) to \(R=4\), unweighted error rises while probability-weighted
  error and \(\Delta\mathrm{NLL}\) fall.  (C)~Across ten models,
  \(D_R(21)=\Delta\mathrm{NLL}_{R=1}-\Delta\mathrm{NLL}_{R=4}\).  Positive values favor finer
  intervals near the row maximum; filled markers have 95\% intervals excluding zero.}
  \label{fig:m-allocation}
\end{figure}
\begin{figure}[t]
  \centering
  \includegraphics[width=\textwidth]{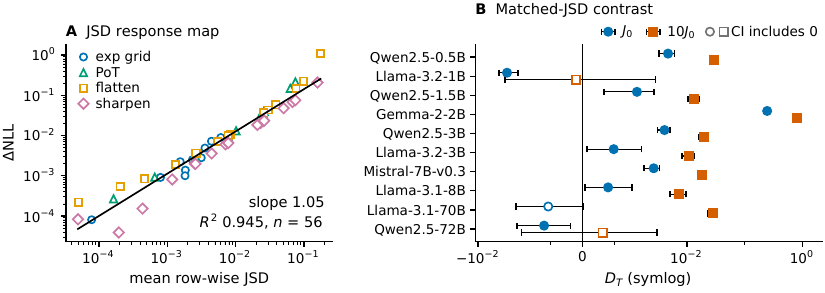}
  \caption{\textbf{Attention divergence and model loss.}  (A)~NLL increase against mean row-wise
  attention JSD for 56 interventions on Qwen2.5-1.5B; the line is an in-sample log--log fit.
  (B)~Flattening-minus-sharpening NLL contrast \(D_T\) after the two perturbations are calibrated
  within each model to the same target mean JSD.  Positive values mean flattening increases NLL
  more than sharpening.  \(J_0\) is each model's JSD under the octave PoT perturbation; circles
  target \(J_0\) and squares target \(10J_0\).  Horizontal bars are 95\% paired-bootstrap
  intervals; hollow markers have intervals including zero.  The signed horizontal axis uses a
  symlog scale.}
  \label{fig:m-structure}
\end{figure}

Allocating more resolution near the row maximum improves model fidelity even when it
increases the overall approximation error.  In Qwen2.5-1.5B at fixed \(K=21\), changing \(R\)
from 1 to 4 increases the unweighted RMS log-weight error by 27.9\%, while it reduces the
probability-weighted error by 29.2\% and \(\Delta\mathrm{NLL}\) by 71.7\%
(Figure~\ref{fig:m-allocation}B).
With only \(R\) varied, \(\Delta\mathrm{NLL}\) falls monotonically from \(0.0161\) at
\(R=0.25\) to \(0.00100\) at \(R=8\) (Figure~\ref{fig:m-allocation}A).

\(D_R(21)\) is positive in all ten model point estimates; nine 95\% intervals exclude zero,
with Gemma-2-2B unresolved (Figure~\ref{fig:m-allocation}C; Table~\ref{tab:crossmodel-summary}).
The interaction \(I_R\) of
Eq.~\eqref{eq:m-dr} is resolved positive in all four tested Qwen2.5 sizes and in
Llama-3.1-70B (Table~\ref{tab:appx-dr}): in these models the allocation benefit is larger at
16 than at 32 intervals.  These comparisons do not establish a scaling relationship, and
\(R=4\) is one tested setting; no optimum was searched.

\subsection{Scalar distortion and model-dependent losses}
\label{sec:m-structure}

Mean row-wise attention JSD tracks the loss caused by an approximation but does not
determine it (Figure~\ref{fig:m-structure}).  Over a fixed set of 56 interventions on
Qwen2.5-1.5B, \(\log\Delta\mathrm{NLL}\) against log mean row-wise JSD has slope \(1.05\) and
\(R^2=0.945\) (in-sample; Figure~\ref{fig:m-structure}A).  Perturbations calibrated to the
same mean row-wise JSD (Eq.~\eqref{eq:m-dt}) nevertheless produce different evaluation
losses.  At \(J_0\), flattening is more damaging than sharpening (\(D_T\) resolved positive)
in seven models, less damaging in Llama-3.2-1B, \(D_T=-0.00313\) \([-0.00428,-0.00202]\), and
Qwen2.5-72B, \(-0.00119\) \([-0.00208,-0.000375]\), and unresolved in Llama-3.1-70B
(Figure~\ref{fig:m-structure}B; Table~\ref{tab:appx-dt}).  At \(10J_0\), \(D_T\) is resolved
positive in eight models and unresolved in Qwen2.5-72B and Llama-3.2-1B.  The Llama-3.2-1B
reversal at \(J_0\) also appears when total variation is matched instead of JSD.  Neither direction is more damaging in all models,
and the pattern does not follow model size.

\label{sec:m-geometry}%
\emph{Similar probability shifts, different losses.}  In all ten models, matched flattening
moves probability out of the set carrying 90\% of the softmax mass and matched sharpening
moves it in, including Llama-3.2-1B, whose contrast is reversed; these leakage summaries have
no intervals.  The change of the local attention output \(pV\) is nearly the same for the two
directions and does not follow the sign of \(D_T\).  Restoring the mass of the two sets is a
diagnostic: it changes the signed contrast in every model at \(10J_0\), but the restored
contrast of Qwen2.5-1.5B is resolved negative, \(-0.00278\) \([-0.00510,-0.000565]\), and
Llama-3.2-1B stays negative (Table~\ref{tab:appx-ggap}).

\subsection{Reconstruction and Rowmax-PoT}
\label{sec:m-recon}

Two further questions concern the grid: how weights are reconstructed from its intervals,
and whether a lattice anchored at the row maximum can replace the full-range grid.  At
\(K=21,R=4\), upper-edge, nearest-boundary and weight-domain interpolation reconstruction
cost \(0.726\%\), \(0.138\%\) and \(0.00808\%\) perplexity (no pairwise intervals).
Interpolation uses the position within the interval, so it does not produce finitely many
output values.  Rowmax-PoT keeps exponential weighting on an octave
lattice anchored at the row maximum: \(\widehat p_j\propto2^{-d_j}\) with
\(d_j=\min\{K_{\max},\lfloor\Delta_j/\ln2+\tfrac12\rfloor\}\).  At \(K_{\max}=20\) it stays below
a 1\% perplexity increase in nine of ten models; Llama-3.1-70B is the exception at
\(1.513\%\) (Table~\ref{tab:crossmodel-summary}).  Section~\ref{sec:m-realization} tests a
kernel implementation.

\section{Rowmax-H15 in a B200 Attention Kernel}
\label{sec:m-realization}

Fidelity and speed are measured separately, under
different dtypes and context lengths.

\begin{figure}[t]
  \centering
  \includegraphics[width=\textwidth]{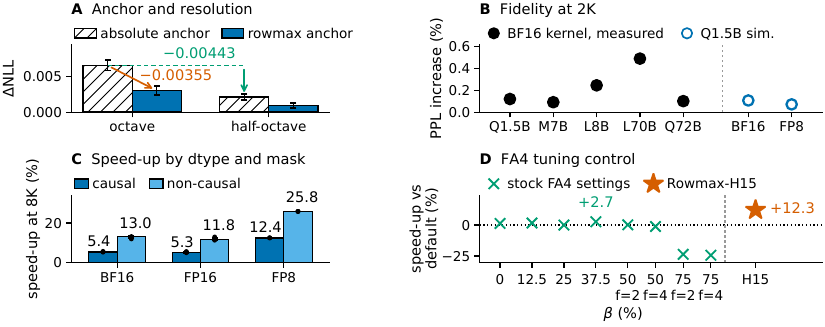}
  \caption{\textbf{Lattice controls, Rowmax-H15 fidelity, and B200 performance.}
  (A)~Qwen2.5-1.5B model evaluations, not B200 kernel measurements.  Orange compares absolute and
  rowmax anchors at octave spacing; green compares octave and half-octave spacing at the absolute
  anchor.  Arrow labels give the resulting decreases in \(\Delta\mathrm{NLL}\); whiskers are 95\%
  paired-bootstrap intervals.  (B)~Perplexity increase over stock FA4 at 2K.  Filled points are
  direct BF16 kernel measurements in five models; hollow points to the right of the divider are
  BF16 and FP8 semantic simulations on Qwen2.5-1.5B.  Model prefixes Q, M, and L denote Qwen2.5,
  Mistral-7B-v0.3, and Llama-3.1, respectively.  Intervals are in Table~\ref{tab:m-h15}; FP8
  kernel fidelity was not measured directly.  (C)~Attention-forward speed-up over stock FA4 at 8K
  by dtype and mask (causal or non-causal).  Bars show speed-ups computed from pooled median call
  latencies; points show measurement rounds and whiskers their standard deviation, not a
  confidence interval.  (D)~Separate FP8 causal 8K sweep against FA4's shipping default.  Green
  crosses vary the share \(\beta\) of polynomially emulated elements and emulation period \(f\);
  the +2.7\% label marks the fastest tested stock setting.  The orange star is Rowmax-H15, which
  has no \(\beta\).  The horizontal dotted line marks zero speed-up; the vertical dashed line
  separates Rowmax-H15 from the stock-setting sweep.}
  \label{fig:m-h15}
\end{figure}

\subsection{Anchor and resolution as separate controls}
\label{sec:m-anchor}

The anchor of a lattice and its spacing can be changed separately, and on Qwen2.5-1.5B both
change the loss (Figure~\ref{fig:m-h15}A; Table~\ref{tab:appx-anchor}).  Starting from an
octave lattice on an absolute scale, as in post-training quantization of probabilities,
moving only the anchor to the row maximum lowers \(\Delta\mathrm{NLL}\) by \(0.00355\)
\([0.00268,0.00448]\), and halving the spacing at the absolute anchor lowers it by
\(0.00443\) \([0.00367,0.00517]\).  The two effects are not additive: the anchor gain is
smaller at half-octave spacing.  Leakage correlates with lattice phase under the absolute
lattices but not under the rowmax octave lattice (Table~\ref{tab:appx-anchor}), which is consistent
with a phase effect but does not demonstrate it.

\subsection{Rowmax-H15}
\label{sec:m-h15}

Rowmax-H15 approximates the base-two exponential by two values per octave.  For the
base-two exponent \(x\) that FA4's online exponential path receives, measured from the
loop's running row maximum,
\begin{equation}
  n=\operatorname{round}_{\mathrm{RNE}}(2x),\qquad
  \widehat w_{\text{Rowmax-H15}}(x)=\left(1+\tfrac12(n\bmod 2)\right)2^{\lfloor n/2\rfloor},
  \label{eq:m-h15}
\end{equation}
so even \(n\) gives \(2^k\) and odd \(n\) gives \(1.5\cdot2^k\)
(Appendix~\ref{sec:realization-h15}); the input is rounded in half-octave steps, and the two
outputs per octave are 0.585 and 0.415 octaves apart.  The anchor is this running maximum, so no scale is calibrated.  That maximum is not always the row's final maximum, so \(x\) can be positive
under FA4's deferred rescale (Appendix~\ref{sec:realization-fidelity}).  At the idealized
operator level Rowmax-H15 coincides with \citeauthor{schraudolph1999fast}'s
\citeyearpar{schraudolph1999fast} unshifted construction evaluated on half-integer-rounded
base-two exponents (Appendix~\ref{app:schraudolph}); in the kernel, one FP32 add and a
22-bit shift compute it (Appendix~\ref{app:patch}).  A static SASS audit of isolated
sm\_86 micro-kernels, not of the B200 kernel, counts 3 instructions (depth 3) per
element against 8 (depth 8) for FA4's cubic emulation.  On Qwen2.5-1.5B, half-octave input
rounding increases offline NLL by 0.00107 [0.000747, 0.00140] relative to the unquantized-input
reference \(S(x)\).  Quarter- and eighth-octave input quantization reduce this loss, but these
variants have not been evaluated inside FA4 (Appendix~\ref{app:expabl}).

\subsection{Kernel experimental setup}
\label{sec:m-hw}

\paragraph{Environment.}  Kernel experiments use one B200 (main
runs: driver 580.126.09, PyTorch 2.14 with CUDA 13.0, FA4 commit \texttt{0dc2cb48}).  We
compare stock FA4 with a patched kernel that changes only the exponential primitive; both
kernels of a comparison run in the same environment, shape and dtype.  The main shape is
batch 1, 16 heads, head dimension 128; Table~\ref{tab:appx-env} lists the environment of
each measurement session.

\paragraph{Latency.}  After 50 warm-up calls, CUDA events time windows of 100 back-to-back
calls; the \emph{call latency}, window time over 100, includes launch gaps.  Three rounds
give each kernel 21 windows; the speed-up is the stock-over-patched ratio of pooled medians minus one, and
the per-round sd describes spread and is not a confidence interval.  A separate Nsight
Systems trace gives the \emph{kernel duration}.

\paragraph{Energy.}  In a 45\,s FP8 causal 16K loop per kernel, \texttt{nvidia-smi} samples board
power at a requested 100\,ms interval; energy per forward is mean power times loop time
over forwards completed.

\subsection{Kernel correctness and model fidelity}
\label{sec:m-fidelity}

\begin{table}[t]
\centering
\footnotesize
\setlength{\tabcolsep}{3.5pt}
\renewcommand{\arraystretch}{0.86}
\caption{\textbf{Rowmax-H15 fidelity, B200 call latency and energy.}  Top: 97 blocks, 95\%
paired-bootstrap intervals; measured rows evaluate the patched kernel directly on the BF16
path; simulation rows are Qwen2.5-1.5B only, and the FP8 row is a semantic simulation, not a
direct FP8 kernel measurement.  Bottom: pooled medians of three rounds; energy from one
sustained loop per kernel.}
\label{tab:m-h15}
\begin{tabular}{@{}lllr@{}}
\toprule
Fidelity path & Model & \(\Delta\)NLL [95\% CI] & PPL change (\%) \\
\midrule
BF16 kernel, measured & Qwen2.5-1.5B & 0.00121 [0.000862, 0.00156] & 0.121 \\
BF16 kernel, measured & Qwen2.5-72B & 0.00101 [0.000665, 0.00139] & 0.101 \\
BF16 kernel, measured & Mistral-7B-v0.3 & 0.000914 [0.000600, 0.00125] & 0.091 \\
BF16 kernel, measured & Llama-3.1-8B & 0.00246 [0.00206, 0.00289] & 0.246 \\
BF16 kernel, measured & Llama-3.1-70B & 0.00491 [0.00402, 0.00582] & 0.492 \\
BF16 semantic simulation & Qwen2.5-1.5B & 0.00107 [0.000680, 0.00145] & 0.107 \\
FP8 semantic simulation & Qwen2.5-1.5B & 0.000716 [0.000370, 0.00106] & 0.072 \\
\midrule
Kernel measurement & Stock FA4 (A0) & Rowmax-H15 (A1) & \(\Delta\) vs stock (\%) \\
\midrule
FP8 causal 8K, hd 128: speed-up & 0.2063 ms & 0.1835 ms & +12.4 \\
FP8 non-causal 8K, hd 128: speed-up & 0.4879 ms & 0.3879 ms & +25.8 \\
FP8 causal 16K: energy per forward & 742.5 mJ & 680.3 mJ & \(-8.4\) \\
\bottomrule
\end{tabular}
\end{table}

We evaluated the patched kernel on the BF16 path in five models from three families, using
97 paired 2K-token blocks and stock FA4 as the baseline.  Perplexity increased by
\(0.091\%\) to \(0.492\%\), and all five 95\% intervals lie above zero; the largest upper bound, Llama-3.1-70B's, is \(0.583\%\) (Table~\ref{tab:m-h15},
Figure~\ref{fig:m-h15}B).  The approximation therefore introduces a small but measurable
loss in model fidelity.  The five models do not establish a trend with model size
(Appendix~\ref{app:fivemodel}).  Element-wise readbacks of the operator are in
Appendix~\ref{app:gates}.

A PyTorch simulation of the kernel semantics, run on Qwen2.5-1.5B only, is a cross-check.  For the BF16 path it gives \(\Delta\mathrm{NLL}=0.00107\), 11\% below the directly measured
BF16-kernel \(\Delta\mathrm{NLL}\), and the two intervals overlap; this agreement is not an equivalence test and does
not validate the FP8 simulation.  The FP8 path was only simulated:
\(\Delta\mathrm{NLL}=0.000716\) (\(+0.072\%\) perplexity), with no resolved difference from
two half-octave reference operators (Table~\ref{tab:appx-h15-full}).  A direct
FP8 run was unusable: per-tensor quantization of the BF16 activations dominated its error (Appendix~\ref{sec:realization-fidelity}).

\subsection{B200 performance and energy}
\label{sec:m-b200}

The patched FP8 causal attention forward (head dimension 128) is 7.1\%, 12.4\% and 12.5\%
faster than stock FA4 in call latency at 4K, 8K and 16K (Table~\ref{tab:m-h15};
Table~\ref{tab:appx-b200}a).  The 4K cell has the largest round-to-round spread (sd 2.13
percentage points); a separate traced session gives \(+9.6\%\) kernel-only at 4K
(Appendix~\ref{app:timeline}).

\emph{dtype \(\times\) mask.}  Measured speed-ups are larger for FP8 and for non-causal
attention.  At causal 8K the speed-up is \(+5.4\%\) at BF16 and \(+5.3\%\) at FP16, which share a
tensor-core rate, and \(+12.4\%\) at FP8, which has twice that rate.  Without the causal mask
the FP8 gain roughly doubles again, to \(+25.8\%\) (all cells in Figure~\ref{fig:m-h15}C).
This is consistent with the softmax work no longer being halved, although the mask also
changes matrix-multiply volume and scheduling.  The FP16 and BF16 non-causal cells were
measured in a later session with more clock variation (Table~\ref{tab:appx-env}).

\emph{FA4's own settings.}  In a separate FP8 causal 8K sweep, Rowmax-H15 is \(12.3\%\) faster
than FA4's default and \(9.3\%\) faster than the fastest of eight stock emulation settings
tested.  That setting and two others were measured in a later session; the sweep does not
establish the optimal stock configuration (Figure~\ref{fig:m-h15}D;
Appendix~\ref{sec:realization-performance}).

\emph{Sequence length.}  Call-latency gains are largest at 8K--16K.  Both measurements show a
slowdown at 2K, while the 1K result changes sign between sessions.  The traced kernel is faster
at every tested length, so shorter kernel duration does not always reduce call latency.  At
32K, the gain falls to \(9.9\%\) with a lower patched-kernel clock (Figure~\ref{fig:seqsweep-supp};
Appendix~\ref{app:timeline}).

\emph{Energy.}  GPU board energy per attention forward falls from 742.5 to 680.3 mJ
(\(-8.4\%\)) at unchanged mean board power, following the shorter loop time per forward.  No
confidence interval was estimated for this single-loop measurement.

\section{Long-Context, Zero-Shot and Vocabulary-Output Checks}
\label{sec:m-breadth}

These checks use the characterization operators of Section~\ref{sec:m-char}, not the
Rowmax-H15 kernel.

\textbf{Long context.}  In the two larger models, \(D_R\) is resolved positive at 2K, 8K and 16K,
and \(R=4\) removes 64.9--78.1\% of the \(R=1\) degradation.  \(D_T\) is resolved only in
Qwen2.5-72B at 2K and 8K, where it is negative; the other four intervals include zero.  Each
context uses its own baseline, \(J_0\) and \(\alpha\) (Appendix~\ref{app:breadth}).

\textbf{Zero-shot.}  On six lm-evaluation-harness tasks the common \(K=32,R=4\) grid changes
the scores of the two larger models by a mean absolute 0.114 percentage points (maximum
0.256), whereas the mean-threshold support control raises LAMBADA perplexity by \(6.727\%\)
in Qwen2.5-72B and \(11.649\%\) in Llama-3.1-70B.  These comparisons are descriptive
(Appendix~\ref{app:breadth}).

\textbf{Vocabulary output.}  The output distribution gives a second measure: in Qwen2.5-1.5B
its JSD against the softmax baseline tracks \(\Delta\mathrm{NLL}\) with Spearman
\(\rho=0.994\) across the default-context conditions, an in-sample association that does not
rule out an endpoint-specific effect.

\section{Limitations}
\label{sec:m-discussion}

\newcommand{\CrossModelSummaryTabular}{%
  \begingroup
  \setlength{\tabcolsep}{4.0pt}%
  \renewcommand{\arraystretch}{0.80}%
  \scriptsize
  \begin{tabular}{@{}lrrrrr@{}}
    \toprule
    Model & \shortstack{Params\\(B)} & \shortstack{$K{=}32,R{=}4$\\PPL inc. (\%)} & \shortstack{$D_R(21)$\\$[95\%\ \mathrm{CI}]$} & \shortstack{Rowmax-PoT $K_{\max}{=}20$\\PPL inc. (\%)} & \shortstack{Rowmax-PoT $K_{\max}{=}20$\\top-1 agree. (\%)} \\
    \midrule
  Qwen2.5-0.5B & 0.49 & 0.134 & 0.00611 [0.00525, 0.00697] & 0.382 & 95.8 \\
  Qwen2.5-1.5B & 1.54 & 0.091 & 0.00351 [0.00282, 0.00420] & 0.310 & 96.2 \\
  Qwen2.5-3B & 3.09 & 0.082 & 0.00331 [0.00256, 0.00404] & 0.289 & 96.4 \\
  Qwen2.5-72B & 72.71 & 0.055 & 0.00196 [0.00137, 0.00258] & 0.252 & 97.6 \\
  Llama-3.2-1B & 1.24 & 0.223 & 0.0109 [0.00976, 0.0120] & 0.950 & 94.4 \\
  Llama-3.2-3B & 3.21 & 0.121 & 0.00629 [0.00540, 0.00714] & 0.704 & 95.2 \\
  Llama-3.1-8B & 8.03 & 0.135 & 0.00462 [0.00383, 0.00542] & 0.531 & 96.0 \\
  Llama-3.1-70B & 70.55 & 0.232 & 0.0101 [0.00833, 0.0119] & 1.513 & 96.1 \\
  Gemma-2-2B & 2.61 & 0.852 & 0.00527 [-0.00369, 0.0132] & 0.353 & 92.1 \\
  Mistral-7B-v0.3 & 7.25 & 0.052 & 0.00167 [0.00119, 0.00218] & 0.235 & 97.5 \\
    \bottomrule
  \end{tabular}%
  \endgroup
}

\begin{table}[t]
  \centering
  \caption{\textbf{Cross-model summary.}  Perplexity increases are point estimates;
  \(D_R(21)\) has model-specific paired-bootstrap intervals; top-1 is the vocabulary-output
  agreement with softmax under Rowmax-PoT.  No interval is pooled across models.}
  \label{tab:crossmodel-summary}
  \CrossModelSummaryTabular
\end{table}

\emph{Model and evaluation scope.}  Results concern frozen inference, primarily WikiText-103 NLL;
zero-shot and vocabulary-output evaluations are supporting checks.  Some effects are unresolved
or reversed in individual models (Sections~\ref{sec:m-allocation}--\ref{sec:m-recon}).

\emph{Kernel quality.}  Kernel fidelity was measured on the BF16 path in five models at 2K,
not at the 8K and 16K timing shapes; FP8 fidelity is a semantic simulation on Qwen2.5-1.5B
only (Appendix~\ref{app:fivemodel}).

\emph{Performance and measurement.}  Timings cover B200 attention forward, not full-model,
decode or backward execution, or B300/GB300 (Appendix~\ref{sec:realization-scope}).  Call-latency
gains are negative at 2K and change sign between sessions at 1K.  At head dimension 64, FP8 gains
are 2.0--3.7\%, but the stock FA4 emulation setting prevents a like-for-like comparison.  Without
performance counters, bottleneck explanations remain inferential.  The pre-registered prediction
of a BF16 gain below 3\% failed: causal 8K gives \(+5.4\%\) (Appendix~\ref{app:predictions}).

\section{Conclusion}
\label{sec:m-conclusion}

Controlled interventions in frozen language models show that support size and grid resolution
can be reduced substantially in the tested settings.  Uniform weighting increases NLL in the
single-layer same-support control on Qwen2.5-1.5B, while finer resolution near the row maximum
lowers NLL at a fixed interval count in all ten model point estimates.  This improvement can
occur despite greater unweighted approximation error, and perturbations calibrated to the same
mean attention JSD can produce different model losses.

Rowmax-H15, motivated by these findings, implements a coarse rowmax-anchored weight
representation inside FlashAttention-4.  On B200, the FP8 attention forward is 12.4--12.5\%
faster in call latency at causal 8K--16K and 25.8\% faster at non-causal 8K.  Separately, BF16
kernel evaluation at 2K increases perplexity by 0.091--0.492\% across five models.  Choosing a
softmax approximation therefore requires measuring how its errors affect the frozen model,
alongside the execution cost of its kernel.

\label{m:end-of-main-text}
\section*{Reproducibility statement}
Appendix~\ref{app:framework} gives the complete intervention definitions, equations,
evaluation protocol, and analysis hierarchy; Appendix~\ref{app:hardware} gives the Rowmax-H15
patch mechanics, correctness gates, latency and energy protocols, the environment of every
session (Table~\ref{tab:appx-env}), the full B200 tables, and the status of the
sealed pre-registered predictions; Appendices~\ref{app:breadth}--\ref{app:structure} give
per-model tables.  All intervals use a 5,000-replicate paired block bootstrap with seed 0,
and the B200 experiments use a fixed FlashAttention-4 commit (Table~\ref{tab:appx-env}).
Across five independent rented instances the kernel smoke test has reproduced the same two
activation means bit-for-bit (Appendix~\ref{app:reproducibility}).  For the four models added
to the fidelity measurement, the stock-kernel NLL on one block was compared with each
model's original run and the token stream was checked to match (Appendix~\ref{app:fivemodel}).  The kernel patch with its apply and revert scripts, the formal results manifest
(models and primary endpoints), the two Nsight Systems summary tables behind Table~\ref{tab:appx-nsys-kernel}, a
provenance file that maps every figure and table to its result files, and the remaining code and
results are available at \CodeURL.

\section*{AI use statement}
Generative AI assistants (code-capable large language models) were used in the design,
implementation, analysis, and writing of this work.  \textbf{Numerical results.}  The reported
attention interventions, model evaluations, bootstrap estimates, kernel correctness checks, B200
latency and energy measurements, figures, and tables derive from executed code and recorded
experimental outputs; no experimental result was supplied by an AI model.  \textbf{Reasoning and
design.}  AI assistants helped discuss hypotheses about attention support, weight resolution, and
row-maximum anchoring; propose controls and paired analyses; draft and revise evaluation,
analysis, kernel-patch, and benchmark code; diagnose experiments; interpret results; examine
related work; and draft, edit, and restructure the manuscript.  \textbf{Mathematical claims.}  AI
assistants were also used to derive and prove the paper's short analytical statements: the reconstruction
and PoT error envelopes (Appendices~\ref{sec:methods-reconstruction} and~\ref{sec:methods-pot}), the
ideal-operator identity \(H(x)=S(Q_{1/2}(x))\) with its octave-error envelope (Appendix~\ref{app:schraudolph}),
and the normalized-error bound used in the correctness gates (Appendix~\ref{app:gates}).  The authors checked
each derivation by hand and against numerical checks on stored outputs; the bit-level tests of
Appendix~\ref{app:expabl} are finite tests, not proofs.  \textbf{Author verification.}  The
authors chose the research questions and experiments, ran or supervised the evaluations and
hardware measurements, reviewed the implementation and reported results against stored outputs,
and reviewed the manuscript and its AI-assisted edits before submission.  The reproducibility
statement and appendices document the correctness checks, measurement protocols, and provenance
of figures and tables.  We take responsibility for the final content of this work, including
code, results, claims, and text produced with the aid of generative AI.

\section*{Ethics statement}
This work involves no human subjects or sensitive data; all models and datasets are
publicly available.

\bibliography{references}

\begin{thebibliography}{30}
\providecommand{\natexlab}[1]{#1}
\providecommand{\url}[1]{\texttt{#1}}
\expandafter\ifx\csname urlstyle\endcsname\relax
  \providecommand{\doi}[1]{doi: #1}\else
  \providecommand{\doi}{doi: \begingroup \urlstyle{rm}\Url}\fi

\bibitem[Chen et~al.(2024)Chen, Liu, Wu, Zheng, Cong, Jiang, Wu, Su, and
  Yang]{int-flashattention-2024}
Shimao Chen, Zirui Liu, Zhiying Wu, Ce~Zheng, Peizhuang Cong, Zihan Jiang,
  Yuhan Wu, Lei Su, and Tong Yang.
\newblock {INT-FlashAttention}: Enabling flash attention for {INT8}
  quantization, 2024.
\newblock URL \url{https://arxiv.org/abs/2409.16997}.

\bibitem[Dao(2024)]{flashattention2-2023}
Tri Dao.
\newblock {FlashAttention-2}: Faster attention with better parallelism and work
  partitioning.
\newblock In \emph{International Conference on Learning Representations
  (ICLR)}, 2024.
\newblock URL \url{https://arxiv.org/abs/2307.08691}.

\bibitem[Dao et~al.(2022)Dao, Fu, Ermon, Rudra, and
  R{\'e}]{flashattention-2022}
Tri Dao, Daniel~Y. Fu, Stefano Ermon, Atri Rudra, and Christopher R{\'e}.
\newblock {FlashAttention}: Fast and memory-efficient exact attention with
  {IO}-awareness.
\newblock In \emph{Advances in Neural Information Processing Systems
  (NeurIPS)}, 2022.
\newblock URL \url{https://arxiv.org/abs/2205.14135}.

\bibitem[Ding et~al.(2022)Ding, Qin, Yan, Chai, Liu, Wei, and
  Liu]{apq-vit-2022}
Yifu Ding, Haotong Qin, Qinghua Yan, Zhenhua Chai, Junjie Liu, Xiaolin Wei, and
  Xianglong Liu.
\newblock Towards accurate post-training quantization for vision transformer.
\newblock In \emph{Proceedings of the 30th ACM International Conference on
  Multimedia}, 2022.

\bibitem[Han et~al.(2026)Han, Wan, Wang, Xie, Yan, You, and
  Zhang]{efq-softmax-2026}
Haohui Han, Yuming Wan, Hongni Wang, Pengcheng Xie, Xiaodong Yan, Runqi You,
  and Wencong Zhang.
\newblock {EFQ-Softmax}: Exp-free quantization for softmax, 2026.
\newblock URL \url{https://arxiv.org/abs/2609.09721}.

\bibitem[\.Islamo\u{g}lu et~al.(2023)\.Islamo\u{g}lu, Scherer, Paulin, Fischer,
  Jung, Garofalo, and Benini]{ita-2023}
Gamze \.Islamo\u{g}lu, Moritz Scherer, Gianna Paulin, Tim Fischer, Victor J.~B.
  Jung, Angelo Garofalo, and Luca Benini.
\newblock {ITA}: An energy-efficient attention and softmax accelerator for
  quantized transformers.
\newblock In \emph{IEEE/ACM International Symposium on Low Power Electronics
  and Design (ISLPED)}, 2023.
\newblock URL \url{https://arxiv.org/abs/2307.03493}.

\bibitem[Kim et~al.(2021)Kim, Gholami, Yao, Mahoney, and Keutzer]{i-bert-2021}
Sehoon Kim, Amir Gholami, Zhewei Yao, Michael~W. Mahoney, and Kurt Keutzer.
\newblock {I-BERT}: Integer-only {BERT} quantization.
\newblock In \emph{International Conference on Machine Learning (ICML)}, 2021.
\newblock URL \url{https://arxiv.org/abs/2101.01321}.

\bibitem[Li \& Gu(2023)Li and Gu]{i-vit-2023}
Zhikai Li and Qingyi Gu.
\newblock {I-ViT}: Integer-only quantization for efficient vision transformer
  inference.
\newblock In \emph{Proceedings of the IEEE/CVF International Conference on
  Computer Vision (ICCV)}, 2023.
\newblock URL \url{https://arxiv.org/abs/2207.01405}.

\bibitem[Li et~al.(2023)Li, Xiao, Yang, and Gu]{repq-vit-2023}
Zhikai Li, Junrui Xiao, Lianwei Yang, and Qingyi Gu.
\newblock {RepQ-ViT}: Scale reparameterization for post-training quantization
  of vision transformers.
\newblock In \emph{Proceedings of the IEEE/CVF International Conference on
  Computer Vision (ICCV)}, 2023.
\newblock URL \url{https://arxiv.org/abs/2212.08254}.

\bibitem[Lin et~al.(2022)Lin, Zhang, Sun, Li, and Zhou]{fq-vit-2022}
Yang Lin, Tianyu Zhang, Peiqin Sun, Zheng Li, and Shuchang Zhou.
\newblock {FQ-ViT}: Post-training quantization for fully quantized vision
  transformer.
\newblock In \emph{Proceedings of the 31st International Joint Conference on
  Artificial Intelligence (IJCAI)}, 2022.
\newblock URL \url{https://arxiv.org/abs/2111.13824}.

\bibitem[Liu et~al.(2024)Liu, Tao, Zou, Chow, Fan, Lei, Pan, Sylvester,
  Kielian, and Saligane]{consmax-2024}
Shiwei Liu, Guanchen Tao, Yifei Zou, Derek Chow, Zichen Fan, Kauna Lei, Bangfei
  Pan, Dennis Sylvester, Gregory Kielian, and Mehdi Saligane.
\newblock {ConSmax}: Hardware-friendly alternative softmax with learnable
  parameters.
\newblock In \emph{IEEE/ACM International Conference on Computer-Aided Design
  (ICCAD)}, 2024.

\bibitem[Lv et~al.(2024)Lv, Chen, Guo, Ding, and Liu]{ptq4sam-2024}
Chengtao Lv, Hong Chen, Jinyang Guo, Yifu Ding, and Xianglong Liu.
\newblock {PTQ4SAM}: Post-training quantization for segment anything.
\newblock In \emph{IEEE/CVF Conference on Computer Vision and Pattern
  Recognition (CVPR)}, 2024.

\bibitem[Ranjan \& Savakis(2024)Ranjan and Savakis]{lrp-qvit-2024}
Navin Ranjan and Andreas Savakis.
\newblock {LRP-QViT}: Mixed-precision vision transformer quantization via
  layer-wise relevance propagation, 2024.
\newblock URL \url{https://arxiv.org/abs/2401.11243}.

\bibitem[Schraudolph(1999)]{schraudolph1999fast}
Nicol~N. Schraudolph.
\newblock A fast, compact approximation of the exponential function.
\newblock \emph{Neural Computation}, 11:\penalty0 853--862, 1999.

\bibitem[Shah et~al.(2024)Shah, Bikshandi, Zhang, Thakkar, Ramani, and
  Dao]{flashattention3-2024}
Jay Shah, Ganesh Bikshandi, Ying Zhang, Vijay Thakkar, Pradeep Ramani, and Tri
  Dao.
\newblock {FlashAttention-3}: Fast and accurate attention with asynchrony and
  low-precision.
\newblock In \emph{Advances in Neural Information Processing Systems
  (NeurIPS)}, 2024.
\newblock URL \url{https://arxiv.org/abs/2407.08608}.

\bibitem[Shankar et~al.(2026)Shankar, Liu, Yang, Xu, Hoehnerbach, Xie, Mohan,
  Xu, Zhu, Fromm, Yu, Leung, and Bocharov]{lp-fa4-mxfp8-2026}
Devashish Shankar, Darren Liu, Chunzhi Yang, Jiaqi Xu, Markus Hoehnerbach,
  Jason Xie, Santosh Mohan, Han Xu, Rich Zhu, Josh Fromm, Hongtao Yu, Max
  Leung, and John Bocharov.
\newblock Low precision {Flash Attention 4}: End-to-end block-scaled attention
  for {Blackwell}.
\newblock PyTorch Blog, September 2026.
\newblock URL
  \url{https://pytorch.org/blog/low-precision-flash-attention-4-end-to-end-block-scaled-attention-for-blackwell/}.

\bibitem[Shi et~al.(2024)Shi, Cheng, Mao, and Wang]{p2-vit-2024}
Huihong Shi, Xin Cheng, Wendong Mao, and Zhongfeng Wang.
\newblock {P$^2$-ViT}: Power-of-two post-training quantization and acceleration
  for fully quantized vision transformer, 2024.
\newblock URL \url{https://arxiv.org/abs/2405.19915}.

\bibitem[Shkolnik et~al.(2024)Shkolnik, Fishman, Chmiel, Ben-Yaacov, Banner,
  and Levy]{exaq-2024}
Moran Shkolnik, Maxim Fishman, Brian Chmiel, Hilla Ben-Yaacov, Ron Banner, and
  Kfir~Yehuda Levy.
\newblock {EXAQ}: Exponent aware quantization for {LLM}s acceleration.
\newblock \emph{arXiv preprint arXiv:2410.03185}, 2024.

\bibitem[Stevens et~al.(2021)Stevens, Venkatesan, Dai, Khailany, and
  Raghunathan]{softermax-2021}
Jacob~R. Stevens, Rangharajan Venkatesan, Steve Dai, Brucek Khailany, and Anand
  Raghunathan.
\newblock Softermax: Hardware/software co-design of an efficient softmax for
  transformers.
\newblock In \emph{58th ACM/IEEE Design Automation Conference (DAC)}, 2021.
\newblock URL \url{https://arxiv.org/abs/2103.09301}.

\bibitem[Sun et~al.(2026)Sun, Li, Zou, Du, Zhang, Dong, Fan, and
  Wang]{vfa-2026}
Yupeng Sun, Yanzhao Li, Zhiqiang Zou, Bai Du, Zhiyuan Zhang, Hui Dong, Gaoyige
  Fan, and Hui Wang.
\newblock {VFA}: Relieving vector operations in flash attention with global
  maximum pre-computation, 2026.
\newblock URL \url{https://arxiv.org/abs/2604.12798}.

\bibitem[Sun et~al.(2025)Sun, Li, Zhang, Pan, Dong, Guo, and
  Wang]{efficient-attention-survey-2025}
Yutao Sun, Zhenyu Li, Yike Zhang, Tengyu Pan, Bowen Dong, Yuyi Guo, and
  Jianyong Wang.
\newblock Efficient attention mechanisms for large language models: A survey,
  2025.
\newblock URL \url{https://arxiv.org/abs/2507.19595}.

\bibitem[Tang et~al.(2024)Tang, Wang, Guo, Tu, Han, Hu, and
  Tao]{transformer-compression-survey-2024}
Yehui Tang, Yunhe Wang, Jianyuan Guo, Zhijun Tu, Kai Han, Hailin Hu, and
  Dacheng Tao.
\newblock A survey on transformer compression, 2024.
\newblock URL \url{https://arxiv.org/abs/2402.05964}.

\bibitem[Xu et~al.(2026)Xu, Chen, Yang, Hoehnerbach, Liu, Zadouri, Shankar,
  Zhou, Yu, Ren, Xu, Yang, Nie, Zhang, Li, Shu, Sultan, Leung, Bocharov, and
  Dao]{gdpa-2026}
Jiaqi Xu, Chao Chen, Shuqi Yang, Markus Hoehnerbach, Xiaoyi Liu, Ted Zadouri,
  Devashish Shankar, Jacky Zhou, Hongtao Yu, Manman Ren, Han Xu, Chunzhi Yang,
  Jade Nie, Haoyu Zhang, Huayu Li, Michael Shu, Musharaf Sultan, Max Leung,
  John Bocharov, and Tri Dao.
\newblock Generalized dot-product attention: Tackling real-world challenges in
  {GPU} training kernels.
\newblock PyTorch Blog, March 2026.
\newblock URL
  \url{https://pytorch.org/blog/generalized-dot-product-attention-tackling-real-world-challenges-in-gpu-training-kernels/}.

\bibitem[Yu et~al.(2026)Yu, Lin, Kong, Chen, Sun, Zeng, Lan, Li, Zheng, Yue,
  Ke, Yi, Hu, Ding, Yao, and Wang]{mxattention-2026}
Jianlin Yu, Jing Lin, Linghui Kong, Aiyue Chen, Weiyi Sun, Chenyu Zeng, Wangli
  Lan, Jinxi Li, Zhuo Zheng, Ziyang Yue, Danning Ke, Fei Yi, Tianchi Hu, Yuan
  Ding, Yiwu Yao, and Junsong Wang.
\newblock {MXAttention}: Data-free optimal scaling and pre-normalization
  quantization for {MXFP4} attention, 2026.
\newblock URL \url{https://arxiv.org/abs/2607.24377}.

\bibitem[Yuan et~al.(2022)Yuan, Xue, Chen, Wu, and Sun]{ptq4vit-2022}
Zhihang Yuan, Chenhao Xue, Yiqi Chen, Qiang Wu, and Guangyu Sun.
\newblock {PTQ4ViT}: Post-training quantization for vision transformers with
  twin uniform quantization.
\newblock In \emph{European Conference on Computer Vision (ECCV)}, 2022.
\newblock URL \url{https://arxiv.org/abs/2111.12293}.

\bibitem[Zadouri et~al.(2026)Zadouri, Hoehnerbach, Shah, Liu, Thakkar, and
  Dao]{flashattention4-2026}
Ted Zadouri, Markus Hoehnerbach, Jay Shah, Timmy Liu, Vijay Thakkar, and Tri
  Dao.
\newblock {FlashAttention-4}: Algorithm and kernel pipelining co-design for
  asymmetric hardware scaling.
\newblock In \emph{Proceedings of Machine Learning and Systems (MLSys)}, 2026.
\newblock URL \url{https://arxiv.org/abs/2603.05451}.

\bibitem[Zhang et~al.(2025{\natexlab{a}})Zhang, Huang, Zhang, Wei, Zhu, and
  Chen]{sageattention2-2024}
Jintao Zhang, Haofeng Huang, Pengle Zhang, Jia Wei, Jun Zhu, and Jianfei Chen.
\newblock {SageAttention2}: Efficient attention with thorough outlier smoothing
  and per-thread {INT4} quantization.
\newblock In \emph{International Conference on Machine Learning (ICML)},
  2025{\natexlab{a}}.
\newblock URL \url{https://arxiv.org/abs/2411.10958}.

\bibitem[Zhang et~al.(2025{\natexlab{b}})Zhang, Wei, Huang, Zhang, Zhu, and
  Chen]{sageattention-2024}
Jintao Zhang, Jia Wei, Haofeng Huang, Pengle Zhang, Jun Zhu, and Jianfei Chen.
\newblock {SageAttention}: Accurate 8-bit attention for plug-and-play inference
  acceleration.
\newblock In \emph{International Conference on Learning Representations
  (ICLR)}, 2025{\natexlab{b}}.
\newblock URL \url{https://arxiv.org/abs/2410.02367}.

\bibitem[Zhang et~al.(2025{\natexlab{c}})Zhang, Wei, Zhang, Xu, Huang, Wang,
  Jiang, Chen, and Zhu]{sageattention3-2025}
Jintao Zhang, Jia Wei, Pengle Zhang, Xiaoming Xu, Haofeng Huang, Haoxu Wang,
  Kai Jiang, Jianfei Chen, and Jun Zhu.
\newblock {SageAttention3}: Microscaling {FP4} attention for inference and an
  exploration of 8-bit training, 2025{\natexlab{c}}.
\newblock URL \url{https://arxiv.org/abs/2505.11594}.

\bibitem[Zhong et~al.(2026)Zhong, Feng, Zhou, Peng, and
  Yu]{zhong2026intattention}
Wanli Zhong, Haibo Feng, Zirui Zhou, Hanyang Peng, and Shiqi Yu.
\newblock {IntAttention}: A fully integer attention pipeline for efficient edge
  inference.
\newblock In \emph{Proceedings of the Ninth Conference on Machine Learning and
  Systems (MLSys)}, 2026.
\newblock arXiv:2511.21513.

\end{thebibliography}
\bibliographystyle{iclr2027_conference}

\clearpage
\appendix
\renewcommand{\thefigure}{S\arabic{figure}}\renewcommand{\theHfigure}{S\arabic{figure}}
\setcounter{figure}{0}
\setlength{\LTpre}{\medskipamount}\setlength{\LTpost}{\medskipamount}%
\FloatBarrier
\section{Operator definitions and evaluation protocol}
\label{app:framework}\label{sec:methods}
\setcounter{table}{0}\renewcommand{\thetable}{A\arabic{table}}\renewcommand{\theHtable}{A\arabic{table}}

The appendices follow the paper: definitions and protocol (A), the Rowmax-H15 kernel and its B200 evaluation (B),
long-context and downstream checks (C), further characterization and mechanism evidence (D), and related work (E).

\FloatBarrier
\subsection{Problem setting and softmax attention}
\label{sec:methods-framework}

All parameters of the pretrained decoder-only models stay frozen; only the map from a row of attention scores to its
normalized weights is replaced, in every head and layer unless stated otherwise.  For one query row, \(\mathcal V\) is
the set of keys permitted by the model's mask, \(r_j=\gamma_{\mathrm{model}}q^\top k_j\) the logit with the model's own
scaling, and \(s_j=g_{\mathrm{model}}(r_j)\) the finite score passed to softmax; \(g_{\mathrm{model}}\) is the identity
except where the architecture specifies an attention-logit transformation (the Gemma softcap is preserved).  With
\begin{equation}
  s_{\max}=\max_{j\in\mathcal V}s_j,\qquad \Delta_j=s_{\max}-s_j,\qquad w_j=\exp(-\Delta_j),\qquad
  p_j=\frac{w_j}{\sum_{k\in\mathcal V}w_k},
  \label{eq:exact-attention}
\end{equation}
masked entries have probability zero.  The \emph{softmax} reference \(p\) is recomputed at each layer from the scores
produced under the current intervention; it is not cached from the unmodified model.

For a positive, full-support approximation (\(0<\widehat w_j<\infty\) for all \(j\in\mathcal V\)) the log-weight error
and the induced distribution are
\begin{equation}
  \epsilon_j=\log \widehat w_j-\log w_j,\qquad
  \widehat p_j=\frac{p_j\exp(\epsilon_j)}{\sum_{k\in\mathcal V}p_k\exp(\epsilon_k)},
  \label{eq:log-weight-error}
\end{equation}
\begin{equation}
  \widehat p_j-p_j\approx p_j\!\left(\epsilon_j-\mathbb E_p[\epsilon]\right)\quad\text{(leading order)}.
  \label{eq:leading-order-attention}
\end{equation}
A common multiplicative error cancels under normalization, whereas variation of \(\epsilon\) across a row changes the
relative weights.  Eq.~\eqref{eq:leading-order-attention} therefore motivates probability-weighted error summaries,
without assuming that they suffice.  It also motivates evaluating the normalized operator in the frozen model instead of
function-level error criteria such as \citeauthor{schraudolph1999fast}'s \citeyearpar{schraudolph1999fast}.  Here \(\epsilon\) is a natural-log error; the octave error \(\varepsilon_2\) of
Appendix~\ref{app:schraudolph} satisfies \(\epsilon=\ln(2)\,\varepsilon_2\).  The six intervention dimensions
(Table~\ref{tab:dof}) are support, within-support weighting, the number of grid intervals, their allocation across a score
row, reconstruction from the grid, and exponential evaluation (the arithmetic that produces the weight itself, changed by Rowmax-PoT and Rowmax-H15).
Eqs.~\eqref{eq:log-weight-error}--\eqref{eq:leading-order-attention} apply only to positive full-support operators;
support-changing interventions are defined by their support distributions and carry no finite \(\epsilon_j\) on
removed entries.

\FloatBarrier
\subsection{Models and evaluation protocol}
\label{sec:methods-evaluation}

The core suite has eight models from four families (Qwen2.5-0.5B, Qwen2.5-1.5B, Qwen2.5-3B, Llama-3.2-1B,
Llama-3.2-3B, Llama-3.1-8B, Gemma-2-2B, Mistral-7B-v0.3); two larger models, Qwen2.5-72B and Llama-3.1-70B, extend the
shared experiments.  Ten-model statements apply only to the ten-model endpoints; the long-context and zero-shot checks
add the two larger models to smaller subsets, and the probes run only on Qwen2.5-1.5B do not include them.  Qwen2.5-1.5B,
the reference model for detailed diagnostic sweeps, receives the full set of diagnostics; the other models receive a core
set with selected extensions, so not every diagnostic was run on every model.

The primary evaluation uses the WikiText-103 raw test split, joined in order, tokenized once per tokenizer, truncated
to 200,000 tokens, cached and verified by a token-sequence signature.  At \(L=2048\) the stream gives 97 aligned,
non-overlapping blocks (the remainder is discarded), each with \(L-1=2047\) next-token predictions, 198,559 in total;
every condition of a model uses the same blocks and block-specific baseline.  Evaluation runs with batch size one, in
evaluation mode, without training and with the KV cache disabled.  Parameters and attention values are BF16;
\(QK^\top\), masking and the score-to-probability computation are FP32, and the probabilities are cast to the value
dtype before the \(pV\) product.  Model-defined attention-logit and final-logit softcaps are retained, and the
language-model head is evaluated in chunks with FP32 logits.  Query chunking (fixed at 512 by a memory gate) does not
change row statistics.  Engineering gates (quantizer mirrors, chunking and mask/GQA equivalence, agreement with the eager path and
the native loss, temperature identity) are passed per model before any task runs; they are prerequisites, not results.  With condition \(c\) and softmax baseline \(0\),
\begin{equation}
  \Delta\operatorname{NLL}_{b,c}=\operatorname{NLL}_{b,c}-\operatorname{NLL}_{b,0},\qquad
  \frac{\operatorname{PPL}_{c}}{\operatorname{PPL}_{0}}=\exp\!\left(\Delta\operatorname{NLL}_{c}\right),
  \label{eq:downstream-endpoints}
\end{equation}
where dataset-level NLL is the prediction-weighted average of block NLLs.  All comparisons are within model; absolute
perplexities are not compared across tokenizers.

\FloatBarrier
\subsection{Support and within-support weighting interventions}
\label{sec:methods-support}

The mean-threshold rule keeps the entries above the row's mean valid score (the row maximum if the set would be empty):
\begin{equation}
  \bar s=\frac{1}{|\mathcal V|}\sum_{j\in\mathcal V}s_j,\quad \mathcal M=\{j\in\mathcal V:s_j>\bar s\},\quad
  p^{\mathrm{rel}}_j=\frac{\exp(s_j)\mathbf 1[j\in\mathcal M]}{\sum_{k\in\mathcal M}\exp(s_k)},\quad
  p^{\mathrm{unif}}_j=\frac{\mathbf 1[j\in\mathcal M]}{|\mathcal M|}.
  \label{eq:mean-support}
\end{equation}
Applied in all layers, the two weightings reach later layers on different trajectories, so the all-layer sweep is not the
weighting control.  The Qwen2.5-1.5B sweep also keeps the largest \(k_i=\lceil r|\mathcal V_i|\rceil\) scores per row
for \(r\in\{0.50,0.25,0.15,0.10,0.065,0.026,0.01,0.0065\}\) under both weightings; support density is the pooled ratio of
selected to valid entries.

\paragraph{Strict single-layer same-support control.}
For each target layer \(\ell\) of Qwen2.5-1.5B, both weightings use softmax below \(\ell\), so they reach \(\ell\) with the same
hidden state, scores, valid mask and deterministic support (Eq.~\eqref{eq:mean-support}); at \(\ell\) they differ only
between \(p^{\mathrm{rel}}\) and \(p^{\mathrm{unif}}\), and above \(\ell\) both return to softmax on their own
activations.  The endpoint is
\begin{equation}
  D_W(\ell)=\mathbb E_b\!\left[\Delta\operatorname{NLL}_{b,\mathrm{unif},\ell}-\Delta\operatorname{NLL}_{b,\mathrm{rel},\ell}\right],
  \label{eq:dw-layer}
\end{equation}
and the fixed-layer aggregate averages the 28 layerwise contrasts within each block before averaging over blocks
(layers are not statistical units).  Same support at the target layer holds by construction; a runtime audit confirms
that selected count, valid count and density agree exactly between the two weightings at every layer (stored Boolean masks are not
compared element by element).

\FloatBarrier
\subsection{Full-range quantization, weighting geometry, and precision allocation}
\label{sec:methods-quantization}

The grid experiments never clip support.  Per row,
\begin{equation}
  s_{\min}=\min_{j\in\mathcal V}s_j,\qquad C=s_{\max}-s_{\min},\qquad u_j=s_j-s_{\min}\in[0,C],
  \label{eq:full-range-coordinate}
\end{equation}
and \([0,C]\) is divided into \(K\) intervals with boundaries \(0=e_0<\cdots<e_K=C\): \(e_a=aC/K\) at \(R=1\), otherwise
\begin{equation}
  \rho=R^{-1/(K-1)},\qquad e_a=C\frac{1-\rho^a}{1-\rho^K},\qquad a=0,\ldots,K,
  \label{eq:allocation-grid}
\end{equation}
so that \(R=h_1/h_K\) with \(h_a=e_a-e_{a-1}\): \(R>1\) makes intervals narrower near the row maximum and \(R<1\) is the
anti-allocation control.  The grid is recomputed for every row.  The weighting-geometry comparison fixes full support,
\(R=1\), upper-edge reconstruction and \(K\), and changes only the map from the upper bin index \(\ell_j\) to the weight:
\(\widehat w_j\propto\exp(e_{\ell_j})\) or the tested linear \(\widehat w_j\propto\ell_j\), both normalized over the
row (Qwen2.5-1.5B: \(K\in\{4,8,12,16,24,32,48,64,128,256,512,1024\}\); other models at \(K=16\)).  The precision
characterization fixes exponential weighting, nearest-boundary reconstruction and \(R=4\), with
\(K\in\{16,32,64,128,256\}\) in Qwen2.5-1.5B and \(K=32\) across models; \(K\) counts intervals, and the reported
\(\log_2(K+1)\) index budget is descriptive, not a storage specification.  The allocation experiment fixes exponential
weighting, full coverage and nearest-boundary reconstruction at \(K=21\), with \(R\in\{0.25,0.5,1,2,4,8\}\) in
Qwen2.5-1.5B and, for every model,
\begin{equation}
  D_R(K)=\mathbb E_b\!\left[\Delta\operatorname{NLL}_{b,K,R=1}-\Delta\operatorname{NLL}_{b,K,R=4}\right],
  \label{eq:dr}
\end{equation}
\(D_R(21)\) being primary.  For the three core Qwen2.5 sizes and both larger models the precision--allocation interaction is
\begin{equation}
  I_R=D_R(16)-D_R(32).
  \label{eq:allocation-interaction}
\end{equation}
For positive full-support operators, Eq.~\eqref{eq:log-weight-error} is pooled over the accepted row--entry pairs
\(\mathcal A\) uniformly or with softmax weights,
\begin{equation}
  E_{\mathrm{RMS}}=\sqrt{\frac{\sum_{(r,j)\in\mathcal A}\epsilon_{rj}^2}{|\mathcal A|}},\qquad
  E_{p,\mathrm{RMS}}=\sqrt{\frac{\sum_{(r,j)\in\mathcal A}p_{rj}\epsilon_{rj}^2}{\sum_{(r,j)\in\mathcal A}p_{rj}}},
  \label{eq:error-summaries}
\end{equation}
on the softmax trajectory, excluding rows with fewer than 64 valid keys or a score span no larger than \(10^{-3}\); the
Qwen2.5-1.5B pass uses all 97 blocks and also records how mass concentrates across probability bands and \(\Delta\).
These are explanatory summaries, not training objectives.

\FloatBarrier
\subsection{Reconstruction from the grid}
\label{sec:methods-reconstruction}

For \(u\in[e_{\mathrm{lo}},e_{\mathrm{hi}}]\), \(h=e_{\mathrm{hi}}-e_{\mathrm{lo}}\), three rules are compared at \(K=21\),
\(R=4\): upper edge (\(\widetilde u=e_{\mathrm{hi}}\)), nearest boundary in score coordinate (ties to the lower
boundary), and LERP of the \emph{weights},
\begin{equation}
  t=\frac{u-e_{\mathrm{lo}}}{h},\quad \widehat w_{\mathrm{lerp}}=(1-t)e^{e_{\mathrm{lo}}}+t\,e^{e_{\mathrm{hi}}},\quad
  \epsilon_{\mathrm{lerp}}=\log\!\left[(1-t)+t\,e^{h}\right]-th .
  \label{eq:lerp-error}
\end{equation}
The idealized local envelopes are \(0\le\epsilon\le h\) (upper edge), \(|\epsilon|\le h/2\) (nearest) and the
exponential-secant error of Eq.~\eqref{eq:lerp-error}, second order in \(h\).  Rowwise shifts cancel on normalization.
LERP is a reconstruction diagnostic: it uses the position of \(u\) within its interval, so it does not produce finitely many
output values.

\FloatBarrier
\subsection{PoT perturbations and Rowmax-PoT}
\label{sec:methods-pot}

The power-of-two perturbation quantizes the rowmax-relative log weight without a tail cutoff,
\begin{equation}
  \widehat w_{m,j}=2^{-\operatorname{round}(m\Delta_j/\ln2)/m},\quad
  \epsilon_{m,j}=\Delta_j-\operatorname{round}(m\Delta_j/\ln2)\frac{\ln2}{m},\quad
  |\epsilon_{m,j}|\le\frac{\ln2}{2m},
  \label{eq:pot}
\end{equation}
the bound holding in exact arithmetic and audited with a recorded FP32 allowance.  The \(m=1\) member defines the model-specific distortion reference level
\begin{equation}
  J_0^{(M)}=\mathbb E_{\mathrm{rows}}\left[\operatorname{JSD}\!\left(p,\widehat p_{\mathrm{PoT},m=1}\right)\right].
  \label{eq:k0}
\end{equation}
Rowmax-PoT is a separate representation with a finite tail index; at the formal \(m=1\) round-to-nearest point,
\begin{equation}
  x_j=\frac{\Delta_j}{\ln2},\qquad d_j=\min\!\left\{K_{\max},\left\lfloor x_j+\tfrac12\right\rfloor\right\},\qquad
  \widehat p_j=\frac{2^{-d_j}}{\sum_{k\in\mathcal V}2^{-d_k}}.
  \label{eq:rowmax-pot}
\end{equation}
The main point clamps the tail at \(K_{\max}=20\) (full support, \(K_{\max}+1\) indices); Qwen2.5-1.5B adds a
\(K_{\max}\) sweep and a tail-drop variant.  Rowmax-PoT (integer octaves, finite tail) is distinct from Rowmax-H15 (two
values per octave inside the online FA4 exponential path, Appendix~\ref{app:hardware}).

\FloatBarrier
\subsection{Matched-JSD and matched-TV perturbations}
\label{sec:methods-matching}

With \(m_j=\tfrac12(p_j+q_j)\),
\begin{equation}
  \operatorname{JSD}(p,q)=\tfrac12\operatorname{KL}(p\|m)+\tfrac12\operatorname{KL}(q\|m),\qquad
  \operatorname{TV}(p,q)=\tfrac12\sum_{j\in\mathcal V}|p_j-q_j|.
  \label{eq:divergences}
\end{equation}
The matching coordinate is the mean row-wise divergence over all included layers, heads and rows, each against the
softmax on the measured condition's own trajectory.  Matching equalizes this aggregate only---not its distribution
over layers, heads or rows, nor the downstream trajectories---so it tests the sufficiency of a scalar summary rather
than isolating direction.  Temperature probes use
\begin{equation}
  p_{\alpha,j}=\frac{\exp(\alpha s_j)}{\sum_{k\in\mathcal V}\exp(\alpha s_k)},
  \label{eq:temperature}
\end{equation}
with flattening \(A^-\) (\(\alpha<1\)) and sharpening \(A^+\) (\(\alpha>1\)); \(\alpha\) multiplies the scores.  These are
controlled probes, not proposed approximations.  Targets are \(J_0^{(M)}\) and \(10J_0^{(M)}\) for every model, plus \(3J_0\) for Qwen2.5-1.5B
(\(J_0=0.00261\)).  Each \(\alpha\) is solved independently by bracketing and bisection on mean row-wise divergence
alone, preferring a relative error of at most 1\% and requiring at most 2\%; the largest achieved error over all 48
matched conditions is \(0.962\%\) (Qwen2.5-1.5B, \(J_0\), flattening).  At the default context, matching uses the
first eight aligned blocks; the values of \(\alpha\) are then frozen and both perturbations are evaluated on all 97 blocks.  Divergence
was not recomputed on the full evaluation set, so equal aggregate JSD holds on the eight calibration blocks, not
claimed on the 97.  Downstream NLL is never available to the solver.  The directional endpoint is
\begin{equation}
  D_T(J)=\mathbb E_b\!\left[\Delta\operatorname{NLL}_{b,A^-(J)}-\Delta\operatorname{NLL}_{b,A^+(J)}\right].
  \label{eq:dt}
\end{equation}
Matched TV repeats the procedure for Qwen2.5-1.5B and Llama-3.2-1B: the TV target is the mean row-wise TV of the
PoT member whose \(m\) the JSD target fixed, and \(D_{TV}\) is defined as in Eq.~\eqref{eq:dt}.

\FloatBarrier
\subsection{High-mass redistribution, local output fidelity, and rescue}
\label{sec:methods-rescue}

For a softmax row \(p\), \(S_q(p)\) is the smallest descending-probability prefix whose mass reaches \(q\) (an entry is
included if the mass before it is below \(q\)); \(q=0.9\) is primary and \(q\in\{0.5,0.8,0.9,0.95\}\) is used in
Qwen2.5-1.5B.  Signed leakage,
\begin{equation}
  L_q=p(S_q)-\widehat p(S_q),
  \label{eq:leakage}
\end{equation}
is positive when mass leaves the high-mass set, which is recomputed at every layer and row on the current trajectory.
For \(y=pV\) and \(\widehat y=\widehat pV\) the local output diagnostic is the pooled relative RMS
\begin{equation}
  E_{pV}=\sqrt{\frac{\sum_r\|\widehat y_r-y_r\|_2^2}{\sum_r\|y_r\|_2^2}}
  \label{eq:pv-rms}
\end{equation}
over accepted rows, heads and layers, with mean row-wise cosine also recorded.  Leakage and \(pV\) use the first four
default-context blocks and rows with at least 64 valid keys; they are descriptive, without bootstrap intervals.

The mass-partition rescue is applied to the matched flattening and sharpening perturbations only.  With \(S=S_{0.9}(p)\),
\(P_{\mathrm{in}}=p(S)\), \(P_{\mathrm{out}}=1-P_{\mathrm{in}}\) and the perturbed \(\widehat P_{\mathrm{in}}\),
\(\widehat P_{\mathrm{out}}\),
\begin{equation}
  \widetilde p_j=
  \begin{cases}
    \widehat p_j\,P_{\mathrm{in}}/\widehat P_{\mathrm{in}}, & j\in S,\\
    \widehat p_j\,P_{\mathrm{out}}/\widehat P_{\mathrm{out}}, & j\notin S,
  \end{cases}
  \label{eq:mass-rescue}
\end{equation}
followed by renormalization: the softmax two-region mass split is restored while each region keeps the perturbed
conditional distribution.  Rescue acts at every layer and valid row, with \(S\) recomputed on the rescued trajectory; it
fails closed if a region collapses and audits the restored partition on rows with at least 64 valid keys.  With
\(D_{\mathrm{orig}}=\Delta\operatorname{NLL}(A^-)-\Delta\operatorname{NLL}(A^+)\) and \(D_{\mathrm{rescue}}\) the same
contrast after rescue,
\begin{equation}
  G_{\mathrm{gap}}=D_{\mathrm{orig}}-D_{\mathrm{rescue}}
  \label{eq:gap-change}
\end{equation}
is a signed algebraic change: \(G_{\mathrm{gap}}>0\) need not mean \(|D_{\mathrm{rescue}}|<|D_{\mathrm{orig}}|\),
particularly when the original contrast is negative or near zero.  No sign is imposed on \(D_T\), \(D_{TV}\) or
\(G_{\mathrm{gap}}\).

\FloatBarrier
\subsection{Analysis hierarchy and uncertainty}
\label{sec:methods-statistics}

The formal results manifest (supplementary material) names five primary endpoints: the strict same-support contrast \(D_W\), \(D_R(21)\),
\(D_T(J_0)\), \(D_T(10J_0)\) and \(G_{\mathrm{gap}}(10J_0)\).  This describes the current analysis hierarchy, not a
historical preregistration.  Supporting analyses are the support and weighting-geometry characterizations, the common
\(K=32\) point, the full \(R\) sweep, \(I_R\) and Rowmax-PoT; secondary mechanism or breadth analyses are reconstruction,
the distortion--response regression, matched TV, alternative high-mass thresholds, leakage, local \(pV\), the \(J_0\)
rescue, vocabulary-output fidelity, the within-context \(D_T\) contrasts and the zero-shot check; the within-context
\(D_R\) contrasts are robustness evidence for the primary allocation conclusion.  The
Rowmax-H15 fidelity and B200 kernel experiments are a downstream realization study with their own scope (measured-kernel
BF16 NLL, FP8 and BF16 semantic simulations, the timing matrix and its controls); they neither redefine the
characterization endpoints nor define a pooled hardware effect.

The statistical unit is the aligned block.  A contrast is \emph{resolved} when its 95\% interval excludes zero and
\emph{unresolved} otherwise; unresolved intervals are always shown.  Condition-level descriptive quantities (support
densities, leakage, \(pV\), zero-shot metrics) carry no intervals, and cross-model results list every model.  For a
blockwise paired contrast \(d_b\) we report the mean and a two-sided 95\% percentile paired-bootstrap interval from
5,000 replicates, each resampling 97 block indices with replacement (seed 0, indices shared across conditions); other
contexts use their available blocks, and \(D_W\) averages layers within block before resampling.  Layerwise \(D_W\)
intervals are pointwise; no interval is adjusted for multiplicity across layers, models or endpoints, no pooled
cross-model effect is computed, and cross-model sign counts are descriptive.

\FloatBarrier
\section{Rowmax-H15 implementation and hardware evaluation}
\label{app:hardware}\label{sec:realization}
\setcounter{table}{0}\renewcommand{\thetable}{B\arabic{table}}\renewcommand{\theHtable}{B\arabic{table}}

\FloatBarrier
\subsection{Representation and implementation}
\label{sec:realization-bridge}
\begingroup
\refstepcounter{table}\label{tab:appx-anchor}
\par\medskip\noindent{\small\textbf{Table~\thetable.} Anchor $\times$ resolution on Qwen2.5-1.5B: 8-bit probability-domain log grids (absolute anchor) versus rowmax-anchored score-domain lattices of the same nominal spacing. Paired block bootstrap over 97 blocks. The 8-bit probability grids are the lattice structure of FQ-ViT (log2) and RepQ-ViT (log$\surd$2), not a reproduction of their 4-bit ViT results.}\par\smallskip
\addcontentsline{lot}{table}{\protect\numberline{\thetable}Anchor $\times$ resolution on Qwen2}
{\scriptsize
\setlength{\tabcolsep}{2pt}
\noindent\begin{tabular}{@{}>{\raggedright\arraybackslash}p{\dimexpr 0.127\linewidth-4pt\relax}>{\raggedright\arraybackslash}p{\dimexpr 0.286\linewidth-4pt\relax}>{\raggedright\arraybackslash}p{\dimexpr 0.276\linewidth-4pt\relax}>{\raggedright\arraybackslash}p{\dimexpr 0.281\linewidth-4pt\relax}@{}}
\toprule
\textbf{lattice spacing} & \textbf{absolute anchor (probability grid)} & \textbf{rowmax anchor (score lattice)} & \textbf{leakage--phase Pearson r (absolute / rowmax)} \\
\midrule
octave & log2-p 8-bit: +0.00657 [+0.00587, +0.00731] & PoT m=1: +0.00303 [+0.00239, +0.00365] & -0.775 / -0.005 \\
half-octave & log$\surd$2-p 8-bit: +0.00214 [+0.00170, +0.00259] & PoT m=2: +0.000927 [+0.000543, +0.00132] & -0.777 / --- \\
\bottomrule
\end{tabular}
}
\par\noindent{\footnotesize Paired differences (absolute reductions in $\Delta$NLL, so a positive value is an improvement; no percentages):}\par
{\scriptsize
\setlength{\tabcolsep}{2pt}
\begin{longtable}{@{}>{\raggedright\arraybackslash}p{\dimexpr 0.574\linewidth-4pt\relax}>{\raggedright\arraybackslash}p{\dimexpr 0.396\linewidth-4pt\relax}@{}}
\toprule
\textbf{change} & \textbf{reduction in $\Delta$NLL [95\% CI]} \\
\midrule
\endhead
anchor: absolute $\rightarrow$ rowmax, octave spacing & 0.00355 [0.00268, 0.00448] \\
anchor: absolute $\rightarrow$ rowmax, half-octave spacing & 0.00122 [0.000708, 0.00171] \\
spacing: octave $\rightarrow$ half-octave, absolute anchor & 0.00443 [0.00367, 0.00517] \\
spacing: octave $\rightarrow$ half-octave, rowmax anchor & 0.00210 [0.00132, 0.00286] \\
\bottomrule
\end{longtable}
\addtocounter{table}{-1}
}
\par\noindent{\footnotesize Conditions: log2-p and log$\surd$2-p round $-\log_2 p$ and $-2\log_2 p$ to 8-bit integers (fixed at $p=1$);
PoT m=1 (m=2) rounds the log weight from the row maximum to octaves (half octaves); absolute-lattice ILPoT uses absolute octaves.
Leakage was not recorded for PoT m=2.  ILPoT-H-A (Table~\ref{tab:appx-h15-full}), not used in this table, is an absolute half-octave
lattice, $\{1,\surd2\}\times2^k$, whose maximum entry is set to its exact weight.}\par
\par\noindent{\footnotesize Ratios of block means: log2-p / PoT m=1 = 2.17; absolute-lattice ILPoT / log2-p = 1.01 (the two operators are algebraically the same up to a per-row phase shift); log$\surd$2-p / PoT m=2 = 2.31; log2-p / PoT m=2 = 7.1.}\par
\endgroup

Rowmax-PoT uses \(e^{-\Delta}=2^{-\Delta/\ln2}\): a \(\log_2\) lattice keeps exponential weighting with powers of two, and the
row-maximum anchor puts each row's largest weight on a lattice point.

\paragraph{Anchor and resolution as separate controls.}
Probability-domain log grids---FQ-ViT's \(\log_2\) and RepQ-ViT's \(\log_{\sqrt2}\)---are score-domain lattices anchored to an
absolute scale, differing from rowmax-anchored ones by a per-row phase (an absolute score-domain octave lattice
reproduces the 8-bit \(\log_2\)-probability grid, ratio 1.01 in \(\Delta\mathrm{NLL}\)).  On Qwen2.5-1.5B
(Table~\ref{tab:appx-anchor}), moving only the anchor at octave spacing lowers \(\Delta\mathrm{NLL}\) by \(0.00355\)
\([0.00268,0.00448]\); halving the spacing at an absolute anchor lowers it by \(0.00443\)
\([0.00367,0.00517]\); the effects are not additive.  Leakage correlates with lattice phase under the absolute
lattices but not under the rowmax octave lattice (it was not recorded for PoT \(m=2\)), consistent with, but not establishing,
a phase mechanism.

\paragraph{Rowmax-H15.}
\label{sec:realization-h15}\label{sec:methods-h15}
With \(x\) the base-two exponent that FA4's online exponential path receives after the running-maximum shift, softmax
scale and configuration offset,
\begin{equation}
  n=\operatorname{round}_{\mathrm{RNE}}(2x),\qquad
  \widehat w_{\text{Rowmax-H15}}(x)=\left(1+\tfrac12(n\bmod2)\right)2^{\lfloor n/2\rfloor},
  \label{eq:h15}
\end{equation}
with round-to-nearest-even, \(\lfloor\cdot\rfloor\) toward \(-\infty\) and \(n\bmod2\) the non-negative residue: even \(n\)
gives \(2^k\), odd \(n\) gives \(1.5\cdot2^k\) (for example, \(n=-3\) gives \(1.5\cdot2^{-2}\)).  Its anchor is the running maximum that online softmax already tracks, so no scale is calibrated; it is exact
at the anchor and phase-aligned while that maximum is current, which is not always the final whole-row maximum
(Appendix~\ref{sec:realization-fidelity}).  With the running maximum current, \(x\le0\) and \(n\) is usually negative; under
FA4's deferred rescale the running maximum can lag, so \(x\) can be positive; Eq.~\eqref{eq:h15} covers both signs.

\paragraph{Implementation.}
\label{app:patch}
Adding \texttt{0x4B400000} (\(2^{23}+2^{22}\); the constant of FA4's floor construction \citep{flashattention4-2026}, here
with round-to-nearest-even) to the doubled FP32 input places \(n\) in the low mantissa bits; a 22-bit left shift moves bit~0
of \(n\) to the leading mantissa position and the rest into the exponent field, and adding the bit pattern of 1.0 supplies
the bias.  Inputs are clamped at \(-252\) (\(-254\) would wrap the exponent).  The patch applies at 10 textual anchors with
fail-closed checks.  Separate exponential-path constants (\(2\cdot\mathtt{scale\_log2}\),
\(2\cdot\mathtt{max\_offset}\)) absorb the doubled input; doubling the shared constants would square the rescale factor,
halve the rescale threshold, corrupt the log-sum-exp and trip the FP8 range assertion.  All other kernel logic (running
maximum, rescaling, log-sum-exp, pipeline, warp roles, FP8 scaling of \(P\)) is unchanged, and the row sum and the \(PV\)
accumulator consume the same approximated weights.  A0 denotes stock FA4 and A1 the patched kernel.

\paragraph{Relation to classical bit-level exponentiation.}
\label{app:schraudolph}
The unshifted base-two idealization of \citeauthor{schraudolph1999fast}'s \citeyearpar{schraudolph1999fast}
construction is \(S(x)=2^{\lfloor x\rfloor}(1+x-\lfloor x\rfloor)\) (offset zero, no finite-mantissa staircase).  With
\(Q_{1/2}(x)=n/2\), \(n=\operatorname{round}_{\mathrm{RNE}}(2x)\), \(S(n/2)\) is \(2^k\) for \(n=2k\) and \(1.5\cdot2^k\) for
\(n=2k+1\), so the ideal operator of Eq.~\eqref{eq:h15} is \(H(x)=S(Q_{1/2}(x))\)---an identity of ideal operators, not
bitwise equivalence between the original macro and our implementation.  With \(q=Q_{1/2}(x)\) the octave error splits into
input rounding and interpolation, \(\varepsilon_2=(q-x)+(\log_2S(q)-q)\), with \(|q-x|\le\tfrac14\) and a second term of
\(0\) at integer and \(\log_2(1.5)-\tfrac12\) at half-integer \(q\); hence \(-\tfrac14\le\varepsilon_2\le\log_2(1.5)-\tfrac14
\approx0.33496\), the \([-0.25,0.335]\) envelope of the checks (closed; not every endpoint need be attained under
ties-to-even).  The half-octave step belongs to the input quantizer; output log-values are alternately \(\log_2(1.5)\) and
\(1-\log_2(1.5)\) apart.  Range, storage rounding, masking and online normalization lie outside the identity.

\FloatBarrier
\subsection{Correctness and model fidelity}
\label{sec:realization-fidelity}
\begin{figure}[!b]
  \centering
  \includegraphics[max width=\textwidth,max height=0.88\textheight]{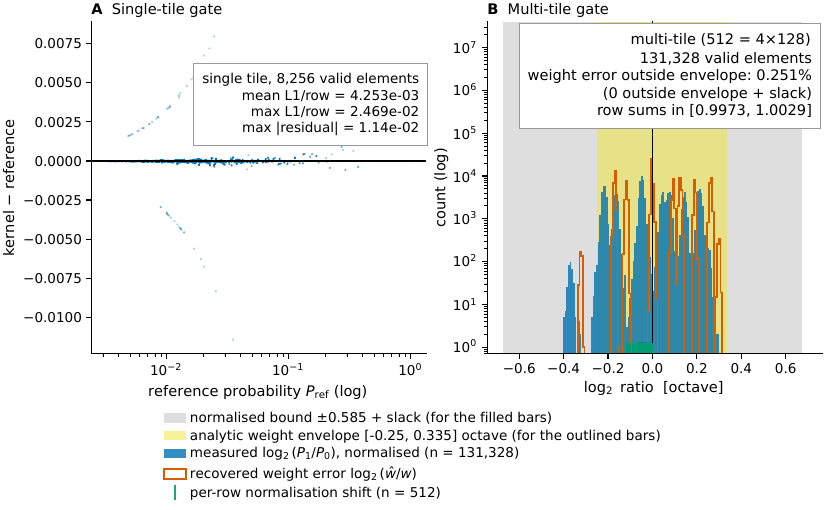}
  \caption{\textbf{Correctness gates for the patched B200 kernel.} Element-wise readbacks measured on B200.  (A)~Single-tile gate: residual of the kernel probabilities against the
  Rowmax-H15 reference over 8,256 valid elements of 128 rows.  (B)~Multi-tile gate over 131,328 valid elements: filled
  bars, the measured normalized ratio \(\log_2(P_1/P_0)\) against the grey \(\pm0.585\)-octave bound it must satisfy;
  outlined bars, the recovered unnormalized error \(\log_2(\hat w/w)\) against the yellow analytic envelope
  \([-0.25,0.335]\); green ticks, the per-row normalization shift read at the row maximum.  Values in
  Appendix~\ref{app:gates}.}
  \label{fig:correctness-gates-supp}
\end{figure}

\paragraph{Correctness gates.}
\label{app:gates}
One-hot value matrices expose the kernel's probabilities (stored readbacks;
Figure~\ref{fig:correctness-gates-supp}).  In the single-tile gate (sequence 128), A1 against a Rowmax-H15 reference built
from the A0 scores gives a mean L1 per row of \(4.253\times10^{-3}\) (maximum \(2.469\times10^{-2}\)) over 128 rows, the size
of E4M3 storage rounding.  The multi-tile gate (sequence 512) reads \emph{normalized} probabilities, whereas the envelope
bounds the \emph{unnormalized} error: since \(\log_2(\hat p_j/p_j)=\epsilon_j-\log_2\sum_k p_k2^{\epsilon_k}\),
\(\epsilon\in[a,b]\) implies only \([a-b,b-a]\) (\(\pm0.585\) octave), which the measured \([-0.3982,0.2923]\) meets once the
E4M3 rounding tolerance (0.09) is allowed for.  At the row maximum, where the operator is exact, the ratio is the per-row normalization shift
(\([-0.1140,0]\), mean \(-0.0495\)); removing it recovers the unnormalized error, \([-0.3324,0.3120]\), with \(0.251\%\) of
131,328 elements below the envelope, none above and none outside it once that tolerance is allowed for.  Row sums lie in \([0.9973,1.0029]\),
and the stock kernel's row-maximum entry attains the patched row maximum in all 512 rows (lattice ties: median 3, at most
38 entries).  A synthetic-input \(O=PV\) error of about 10\% reflects narrow synthetic score ranges.  The checks are not a
correctness proof and do not replace model-level NLL.

\begingroup
\needspace{12\baselineskip}%
\refstepcounter{table}\phantomsection\label{tab:appx-h15-full}%
\par\medskip\noindent{\small\textbf{Table~\thetable.} Model-level fidelity of the Rowmax-H15 operator on Qwen2.5-1.5B (WikiText-103, 97 aligned 2048-token blocks; paired block bootstrap, 5000 replicates, seed 0).}\par\smallskip
\addcontentsline{lot}{table}{\protect\numberline{\thetable}Model-level fidelity of the Rowmax-H15 operator on Qwen2}
{\scriptsize
\setlength{\tabcolsep}{3pt}
\begin{longtable}{@{}>{\raggedright\arraybackslash}p{0.34\linewidth}ll>{\raggedright\arraybackslash}p{0.22\linewidth}@{}}
\toprule
\textbf{Measurement (semantics)} & \textbf{$\Delta$NLL [95\% CI]} & \textbf{PPL change} & \textbf{Evidence status} \\
\midrule
\endhead
B200 patched FlashAttention-4 kernel (A1 $-$ A0) (BF16 path, \texttt{rescale\_threshold = 8.0}) & +0.00121 [0.000862, 0.00156] & +0.121\% & \textbf{direct kernel measurement} \\
Semantic simulation of the BF16 kernel path (BF16 semantics, threshold 8.0) & +0.00107 [+0.000680, +0.00145] & +0.107\% & cross-check of the simulator against the measured kernel \\
Semantic simulation of the FP8 kernel path (FP8 semantics, per-tile running max (tile 128), threshold 0) & +0.000716 [+0.000370, +0.00106] & +0.072\% & \textbf{semantic simulation --- not a direct FP8 kernel measurement} \\
Offline whole-row Rowmax-H15 (control) (rowmax anchor, no tiling) & +0.00105 [+0.000712, +0.00139] & +0.105\% & semantic control \\
Absolute-anchored \(\{1,1.5\}\times2^k\) lattice (control; condition \texttt{ilpot\_h15\_v1}) (absolute anchor) & +0.00224 [+0.00179, +0.00266] & +0.224\% & semantic control (phase-misaligned) \\
PoT m = 2 (half-octave reference) (rowmax anchor) & +0.000927 [+0.000543, +0.00132] & +0.093\% & reference \\
ILPoT-H-A (half-octave reference) (absolute lattice, exact maximum entry) & +0.000859 [+0.000505, +0.00120] & +0.086\% & reference \\
\bottomrule
\end{longtable}
\addtocounter{table}{-1}
}
\par\noindent{\footnotesize Paired differences (block-paired bootstrap):}\par
{\scriptsize
\setlength{\tabcolsep}{2pt}
\begin{longtable}{@{}>{\raggedright\arraybackslash}p{0.32\linewidth}l>{\raggedright\arraybackslash}p{0.34\linewidth}@{}}
\toprule
\textbf{Contrast} & \textbf{$\Delta$($\Delta$NLL) [95\% CI]} & \textbf{Reading} \\
\midrule
\endhead
FP8 semantic Rowmax-H15 $-$ PoT m = 2 & -0.000211 [-0.000612, +0.000183] & No resolved difference at the reported precision; the paired 95\% interval includes zero. \\
FP8 semantic Rowmax-H15 $-$ ILPoT-H-A & -0.000143 [-0.000590, +0.000304] & No resolved difference at the reported precision; the paired 95\% interval includes zero. \\
FP8 semantic Rowmax-H15 $-$ offline whole-row Rowmax-H15 & -0.000333 [-0.000648, -0.0000278] & per-tile running-max semantics differ from whole-row rowmax \\
BF16 semantic Rowmax-H15 $-$ FP8 semantic Rowmax-H15 & +0.000358 [-0.00000769, +0.000736] & Higher point estimate under BF16 conditional-rescale semantics; the paired 95\% interval includes zero. \\
Simulation vs measured kernel (BF16 path) & measured +0.00121 vs simulated +0.00107 & relative gap 11\%; intervals overlap \\
\bottomrule
\end{longtable}
\addtocounter{table}{-1}
}
\par\noindent{\footnotesize Top-1 next-token agreement with the softmax baseline for the FP8 semantic operator: 0.9748. Leakage--phase Pearson r for the FP8 semantic operator: -0.0037 (phase-aligned; compare -0.751 for the absolute octave lattice).}\par
\endgroup

\paragraph{Semantic simulation and the FP8 estimate (Qwen2.5-1.5B only).}
A PyTorch simulation reproduces the kernel semantics in floating point: the per-tile running maximum (tile 128), the
rescale rule and Rowmax-H15 quantization.  It does not quantize \(Q\), \(K\), \(V\) and \(P\) to FP8, and it decides the
rescale per row instead of per warp of 32 rows (Table~\ref{tab:appx-h15-full}).  On the BF16 path it lands 11\% below the measured kernel with
overlapping intervals, a cross-check rather than an equivalence test.  The simulated FP8 path raises perplexity by \(0.072\%\);
\textbf{this is a semantic simulation, not a direct FP8 kernel measurement}, and the BF16 cross-check does not transfer to
it.  A direct FP8 run was not interpretable because per-tensor E4M3 conversion of the BF16 activations added about 0.49
nats of unrelated error.  The FP8 estimate shows no resolved difference from two half-octave references: PoT \(m=2\), anchored at the row
maximum (Eq.~\eqref{eq:pot}), and ILPoT-H-A, an absolute lattice whose maximum entry is exact (notes to Table~\ref{tab:appx-anchor}).

\paragraph{Conditional rescaling moves the anchor.}
FA4 skips the online rescale while the row maximum grows by less than \(\tau=8.0\), so the row's largest weight can leave
the lattice (39.5\% of synthetic rows under the BF16 semantics, none under the FP8 per-tile semantics).  The point estimate
is \(1.50\times\) larger under the BF16 semantics, but the paired difference includes zero.

\paragraph{Exponential-map ablation (offline whole-row path).}
\label{app:expabl}
A separate follow-up run, a supporting analysis outside the primary endpoints of Appendix~\ref{sec:methods-statistics},
on one RTX 3090 (PyTorch 2.5.1, CUDA 12.1) evaluates Qwen2.5-1.5B on the same 97 blocks with the same evaluation code and
bootstrap, and changes only the map from the whole-row base-two exponent \(x\le0\) to a weight: softmax, offline whole-row
Rowmax-H15 \(H(x)=S(Q_{1/2}(x))\), PoT \(m=2\) (its existing log-domain implementation), the unshifted \(S(x)\) of
Appendix~\ref{app:schraudolph} and \(S(Q_{2^{-k}}(x))\) for \(k=2,3\), with
\(Q_{2^{-k}}(x)=2^{-k}\operatorname{round}_{\mathrm{RNE}}(2^kx)\).  Here \(k\) sets the quantization step of the base-two
exponent input; Rowmax-H15 corresponds to \(k=1\).  \(S(x)\) and \(S(Q_{2^{-k}}(x))\) are applied to \(x\) after
Rowmax-H15's lower clamp, \(\max(x,-126)\).  Scores (FP32), anchor, row-sum normalization and query chunking are shared.
The Rowmax-H15 and PoT \(m=2\) rows reproduce the per-block NLLs behind their Table~\ref{tab:appx-h15-full} entries
(largest difference \(4.4\times10^{-16}\)).  All contrasts use aligned per-block NLLs under the same evaluation protocol.
The \(k=2\) and \(k=3\) rows come from a second run whose token sequences and per-block softmax baseline NLLs matched the
first run exactly.
\par\noindent{\footnotesize Exponential-map ablation, same environment (Qwen2.5-1.5B, offline whole-row path, 97 blocks):}\par
{\scriptsize
\setlength{\tabcolsep}{3pt}
\begin{longtable}{@{}>{\raggedright\arraybackslash}p{0.25\linewidth}lll@{}}
\toprule
\textbf{Offline operator} & \textbf{$\Delta$NLL vs softmax [95\% CI]} & \textbf{PPL change} & \textbf{Paired $\Delta$NLL vs Rowmax-H15 [95\% CI]} \\
\midrule
\endhead
Rowmax-H15, \(H(x)=S(Q_{1/2}(x))\) & +0.00105 [+0.000712, +0.00139] & +0.105\% & reference \\
Rowmax-PoT \(m=2\) (Eq.~\eqref{eq:pot}) & +0.000927 [+0.000543, +0.00132] & +0.093\% & \(-0.000122\) [\(-0.000551\), +0.000301] \\
Unshifted Schraudolph \(S(x)\) & \(-0.0000231\) [\(-0.000306\), +0.000270] & \(-0.00231\%\) & \(-0.00107\) [\(-0.00140\), \(-0.000747\)] \\
\(S(Q_{2^{-2}}(x))\), \(k=2\) & +0.000329 [+0.0000270, +0.000624] & +0.033\% & \(-0.000721\) [\(-0.00106\), \(-0.000393\)] \\
\(S(Q_{2^{-3}}(x))\), \(k=3\) & +0.0000796 [\(-0.000156\), +0.000316] & +0.00796\% & \(-0.000970\) [\(-0.00129\), \(-0.000649\)] \\
\bottomrule
\end{longtable}
\addtocounter{table}{-1}
}

\noindent \(S(x)\) shows no resolved change against softmax (the interval includes zero, which does not establish
equivalence) and has lower NLL than Rowmax-H15 on 70 of the 97 blocks: on this path the model-level cost of the half-octave
input rounding is \(0.00107\) \([0.000747, 0.00140]\), whereas the PoT \(m=2\) versus Rowmax-H15 contrast is not
resolved.  These are offline operator NLLs, not kernel measurements, and say nothing about latency, energy or
special-function load.  \(S(x)\) is the unquantized-input fidelity reference for Rowmax-H15; it was not implemented in the
kernel.  Bit-level tests matched the reference \(S(Q_{2^{-k}}(x))\) on the tested inputs for each \(k=0,\ldots,13\), using
the same add-and-shift operation pattern.  Each test compared, bit for bit, the FP32 pattern of Appendix~\ref{app:patch}
with \(2^k\) in place of the doubling and a left shift by \(23-k\) (addition with round-to-nearest-even) against the
reference on the same 2,295,040 inputs in \([-126,0]\): 2,000,000 uniform draws, the integers \(0,\ldots,-126\) and every
multiple of \(7\cdot2^{-14}\), which include round-to-nearest-even ties for every \(k\).  Both sides take the input after
the clamp at \(-126\); no positive \(x\) was tested, and the pattern describes the test's operations, not compiled B200
instructions, register use or latency.  Rowmax-H15 is the \(k=1\) member of this family.  The \(k=2\) and \(k=3\)
variants reduce offline NLL relative to \(k=1\).  Their fidelity and performance inside FA4 remain unmeasured; the
reported kernel results apply only to Rowmax-H15 (\(k=1\)).

\subsubsection{Measured kernel fidelity across five models}
\label{app:fivemodel}
The fifth session measured Qwen2.5-72B, Mistral-7B-v0.3, Llama-3.1-8B and Llama-3.1-70B on the same 97 blocks under the
Qwen2.5-1.5B protocol (BF16 path, conditional rescale in force); the Qwen2.5-1.5B row is the first session's measurement.
The harness requires head dimension 128 and no softcapping.  Llama-3.1-8B and Mistral-7B-v0.3 were fixed in advance;
Llama-3.1-70B and then Qwen2.5-72B were added later, each under a reporting rule fixed before its run.  All five use
grouped-query attention (query/key-value heads 12/2, 64/8, 32/8, 32/8, 64/8), and token identity with the original runs
was checked.  Block~0 under stock FA4 differs from the original run (other hardware and implementation) by \(-0.00136\), \(+0.00130\), \(+0.00036\), \(-0.00014\) and
\(+0.00119\) (Qwen2.5-1.5B, Qwen2.5-72B, Mistral-7B-v0.3, Llama-3.1-8B, Llama-3.1-70B); this wiring check is not an error
bound (\(|-0.00136|\) exceeds the smallest effect, \(0.000914\)) and does not enter the paired contrast.
Table~\ref{tab:m-h15} gives the five estimates and intervals, all above zero; the patched kernel has the lower NLL on 24, 31,
29, 11 and 11 of the 97 blocks (same order).  No pooled estimate is formed.

\subsection{Measurement protocol and environment}
\label{app:reproducibility}\label{app:benchmark}
\begin{table}[!htbp]
\refstepcounter{table}\phantomsection\label{tab:appx-env}%
\noindent{\small\textbf{Table~\thetable.} Environments of the characterization runs and of the five B200
sessions, each B200 session on its own rented instance.  ``Clock and power record'' states what was logged; no session
records a locked SM clock.}\par\smallskip
\addcontentsline{lot}{table}{\protect\numberline{\thetable}Environments of the characterization runs and the B200 sessions}
{\scriptsize\setlength{\tabcolsep}{3pt}
\noindent\begin{tabular}{@{}>{\raggedright\arraybackslash}p{1.35cm}>{\raggedright\arraybackslash}p{4.05cm}>{\raggedright\arraybackslash}p{1.75cm}l>{\raggedright\arraybackslash}p{1.75cm}>{\raggedright\arraybackslash}p{2.35cm}@{}}
\toprule
\textbf{Run group} & \textbf{Measurements} & \textbf{GPU used} & \textbf{Driver} & \textbf{PyTorch / CUDA} & \textbf{Clock and power record} \\
\midrule
Character\-ization, core & Section~\ref{sec:m-char} interventions on the core models; anchor controls and semantic
  simulations (Qwen2.5-1.5B) & one GPU, model not recorded & not recorded & 2.5.1+cu121 / 12.1 & not recorded \\
Character\-ization, large & Section~\ref{sec:m-char} interventions on Llama-3.1-70B and Qwen2.5-72B &
  three A100-SXM4-80GB & 580.173.02 & 2.5.1+cu121 / 12.1 & not recorded \\
B200 S1 & patch and gates; Qwen2.5-1.5B BF16 fidelity; main matrix; sequence sweep and 1K/2K retest; FP8 non-causal
  control; head, batch and head-dimension-64 controls; five tuning settings and Rowmax-H15; energy loop; cuDNN and SDPA
  references; \texttt{ncu} refused & one B200 & 580.126.09 & 2.14.0+cu130 / 13.0 & lock not permitted; clock,
  temperature and power after each window; energy loop sampled \\
B200 S2 & gates re-run with element dumps; GQA (three rounds); \texttt{ncu} refused & one B200 &
  580.126.09 & 2.14.0+cu130 / 13.0 & after each window (GQA at 1965 MHz throughout) \\
B200 S3 & Nsight Systems timeline (kernel duration, call latency, launch cost); \texttt{ncu} refused &
  one B200 & 595.91.07 & 2.14.0+cu130 / 13.0\(^{a}\) & none \\
B200 S4 & FP16 causal and FP16, BF16 non-causal (three rounds); three tuning settings & one B200 &
  580.126.09 & 2.14.0 / 13.0\(^{b}\) & after each window; 1485--1965 MHz and 258--546 W in the three-round cells \\
B200 S5 & BF16 kernel fidelity (measured), four models & one B200 & 580.126.09 & 2.11.0+cu128 / 12.8 &
  no timing \\
\bottomrule
\end{tabular}
}
\par\noindent{\scriptsize The GPU column counts the GPUs a run used; each B200 record lists one visible device.  The
characterization runs keep BF16 weights and compute attention scores and the score-to-probability map in FP32.  FA4 commit
\texttt{0dc2cb48e484894c48a6cc5c6503fa5cded350c3} is recorded for S1--S4 and named in the S5 log; within every A0/A1
comparison both kernels share it.  The \texttt{nvidia-cutlass-dsl} package is 4.7.1 in S1, S2 and S4 and 4.8.0 in S5.
\(^{a}\)S3 left no package snapshot; its versions come from its log, and its trace records the host CPU (Intel Xeon
Platinum 8559C) and OS (Ubuntu 24.04.4 LTS), which no other session records.  \(^{b}\)The S4 package list gives no
PyTorch build tag; its CUDA runtime package is 13.0, as in S1.  Host memory is recorded for no session.\par}
\end{table}

The B200 work comprises five sessions, each on its own rented single-GPU instance (Table~\ref{tab:appx-env}).  Every
A0/A1 comparison is made within one session (same instance, environment, shape and dtype); absolute latencies are not
compared across sessions, and the one cross-session comparison (stock-FA4 sweep) is marked.  A fixed-input smoke test gave the same two
activation means (0.188664 for A0, 0.188038 for A1) bit-for-bit in all five sessions.  No session locked the SM clock (the
first session's attempt was refused); it is read after each window.  In the main matrix it was 1965 MHz
for both kernels in every FP8 cell up to 8K and for A0 at FP8 16K (A1: median 1950 MHz); BF16 held 1965 MHz at 4K, dipped
in some 8K windows and ran at medians of 1672 and 1687 MHz (A0/A1) at 16K.

\paragraph{Call latency.}
Each kernel runs in its own process, because the selection flag is read when the kernel is built.  After fixed-seed inputs, 50
warm-up calls and a synchronization, windows of 100 back-to-back calls of the FA4 Python interface are timed between two
CUDA events, followed by a synchronization; window time over 100 is the call latency, which includes launch and dispatch
gaps but not input generation or compilation.  The speed-up is \(100\,(\tilde t_{A0}/\tilde t_{A1}-1)\) with \(\tilde t\) the
median over a kernel's pooled windows; the sd is the population standard deviation of the per-round speed-ups, a spread,
not a confidence interval.  Unless noted, a cell has three rounds of seven windows per kernel in alternating order (A0A1,
A1A0, A0A1): the main matrix, the sequence sweep, GQA (same structure; its script was not kept) and the fourth session's 8K
FP16 causal and FP16/BF16 non-causal cells.  Exceptions: the FP8 non-causal 8K control ran A0 before A1 in all three
rounds (all 42 windows at 1965 MHz); the 1K/2K retest is one pair of nine windows per length; the head-count, batch-4,
head-dimension-64, 2K/32K-extension and other single-round cells are one pair of seven windows; the stock-FA4 sweep has
seven windows per setting (one A0 process per setting, one A1 process); and the cuDNN and SDPA references have seven
windows per backend without clock record.

\paragraph{Kernel duration.}
The third session traced each kernel and length in its own process with Nsight Systems 2024.6.2 (\texttt{nsys
--trace=cuda}); its call latency is the host time per call over the 200 timed calls, and the kernel duration is the median
of the \texttt{cuda\_gpu\_kern\_sum} report over all 250 traced calls, warm-up included.  Using only the 200 timed calls
changes the kernel-only speed-up by at most 0.1 points from 1K to 8K and at non-causal 8K, but gives \(+10.5\%\) instead of
\(+11.1\%\) at 16K and \(+11.2\%\) instead of \(+9.5\%\) at 32K; Table~\ref{tab:appx-nsys-kernel} keeps the recorded medians.
No headline latency comes from Nsight Systems.

\paragraph{Energy.}
The energy loop ran once per kernel in the first session, A0 then A1: after 30 warm-up calls, \texttt{nvidia-smi} samples
\texttt{power.draw}, \texttt{clocks.sm} and \texttt{temperature.gpu} (requested every 100 ms, not a sensor update period)
from 2 s before to 1 s after a 45 s loop of FP8 causal 16K forwards in chunks of 50 calls, each followed by a
synchronization.  With host loop time \(T\), \(N\) completed calls and \(\bar P\) the mean of the 480 samples minus the first
20 and last 10, \(E_{\mathrm{forward}}=\bar P\,T/N\), with no idle power subtracted.  The first ten retained samples are the
start-up ramp (below 900 W); excluding them raises both energies by about 1.0\% and gives \(-8.36\%\) against the reported
\(-8.4\%\).  The retained samples have median SM clock 1897 MHz (A0) and 1830 MHz (A1), 85\% and 97\% at or above 990 W,
and maxima of 1001.2 W and 1005.4 W; the power limit was not recorded, so no power-capped regime is claimed.  The quantity
is GPU board energy per fused attention forward.

\FloatBarrier
\subsection{Performance and controls}
\label{sec:realization-performance}
\begin{figure}[!b]
  \centering
  \includegraphics[max width=\textwidth,max height=0.88\textheight]{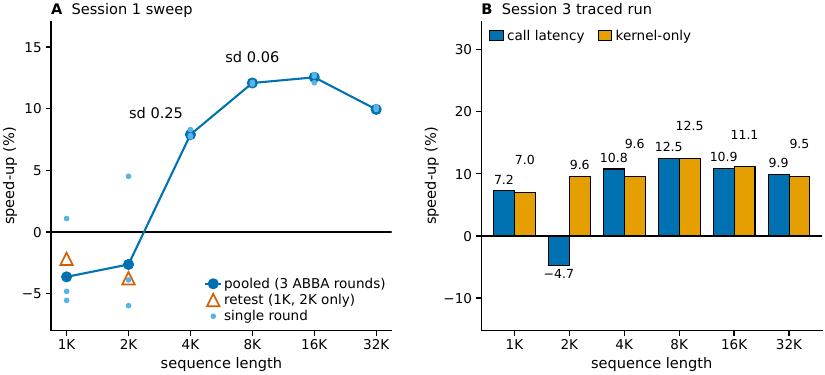}
  \caption{\textbf{Sequence length: the sweep and the traced run, kept apart.} FP8 causal attention-forward speed-up of A1 over A0 (head dimension 128).  (A)~First-session sweep in call latency:
  pooled median over three alternating rounds (line), the rounds (small dots, sd annotated at 4K and 8K) and the separate
  higher-precision 1K/2K retest (triangles; its \(-2.2\%\) and \(-3.7\%\) are the values quoted in the text, the pooled
  sweep gives \(-3.6\%\) and \(-2.6\%\)).  (B)~Third-session traced run: host-timed call latency and kernel-only speed-up
  from the same run.  The panels are different sessions and are not compared across panels.}
  \label{fig:seqsweep-supp}
\end{figure}
\begingroup
\needspace{12\baselineskip}
\refstepcounter{table}\label{tab:appx-nsys-kernel}\label{tab:appx-nsys-launch}
\par\medskip\noindent{\small\textbf{Table~\thetable.} Session-3 timeline (FP8, head dimension 128, 16 heads, batch 1): median kernel duration over all 250 traced calls and host-side call latency over the 200 timed calls, per kernel, with the resulting kernel-only and call-latency speed-ups of A1 over A0 and the share of the timed interval outside the kernel (100 minus the kernel share; launch and host work overlap kernel execution, so the share is not an additive time budget).}\par\smallskip
\addcontentsline{lot}{table}{\protect\numberline{\thetable}Session-3 timeline: kernel duration and call latency}
{\scriptsize\setlength{\tabcolsep}{3pt}
\begin{longtable}{@{}llrrrrrrl@{}}
\toprule
\textbf{seq} & \textbf{mask} & \multicolumn{2}{c}{\textbf{kernel (\(\mu\)s) A0 / A1}} & \multicolumn{2}{c}{\textbf{call latency (\(\mu\)s) A0 / A1}} & \textbf{kernel-only} & \textbf{call latency} & \textbf{non-kernel \% (A0 / A1)} \\
\midrule
\endhead
1K & causal & 17.840 & 16.672 & 45.611 & 42.529 & +7.0\% & +7.2\% & 60.9 / 60.8 \\
2K & causal & 30.688 & 28.000 & 42.916 & 45.027 & +9.6\% & -4.7\% & 28.5 / 37.8 \\
4K & causal & 57.280 & 52.256 & 58.495 & 52.803 & +9.6\% & +10.8\% & 2.1 / 1.0 \\
8K & causal & 200.640 & 178.368 & 201.617 & 179.274 & +12.5\% & +12.5\% & 0.5 / 0.5 \\
16K & causal & 725.920 & 653.296 & 726.815 & 655.539 & +11.1\% & +10.9\% & 0.1 / 0.3 \\
32K & causal & 2747.570 & 2508.674 & 2799.928 & 2548.133 & +9.5\% & +9.9\% & 1.9 / 1.5 \\
8K & non-causal & 487.792 & 379.840 & 488.962 & 380.794 & +28.4\% & +28.4\% & 0.2 / 0.3 \\
\bottomrule
\end{longtable}
\addtocounter{table}{-1}
}
\endgroup

Table~\ref{tab:appx-b200} lists every cell.  The 4K main-matrix cell has the largest round-to-round spread; a separate
traced session shows a positive kernel-duration gain there but does not identify the source of that variability, and 8K
and 16K are the primary cells.

\paragraph{Sequence length.}
\label{app:timeline}
The sweep's call-latency gain (Table~\ref{tab:appx-b200}b) is negative at 1K and 2K (\(-2.2\%\), \(-3.7\%\) in the retest),
peaks at 8K--16K and falls to \(+9.9\%\) at 32K, where A1 ran at a lower clock.  In the traced run
(Table~\ref{tab:appx-nsys-kernel}; Figure~\ref{fig:seqsweep-supp}B) the kernel-only gain is positive at every length, while
the share of the timed interval outside the kernel is large at 1K and 2K and at most 2.1\% from 4K.  From 2K to 4K the
kernel-only gain stays at \(+9.6\%\) while the call-latency gain moves from \(-4.7\%\) to \(+10.8\%\): the operator is the
same, the composition of the timed interval changes.  The \texttt{cudaLaunchKernelExC} call takes 3.9--4.9~\(\mu\)s at all
14 traced points; the kernel, the launch call and an unattributed remainder (expected: Python and CuTe-DSL dispatch) take
39.1/39.2\%, 10.1/10.4\% and 50.8/50.4\% of the timed interval at 1K and 71.5/62.2\%, 10.2/10.0\% and 18.3/27.8\% at 2K
(A0/A1); from 4K the launch overlaps kernel execution and no split is made.  At 2K the non-kernel residual (call latency
minus kernel duration) differs between the kernels by 4.8~\(\mu\)s, as large as the gap between the two gains, but as a residual of
overlapping measurements it is not a measured host time.  The gains agree at 8K and 16K, which checks the traced session
only, not the first session's CUDA-event timing.  The 1K call-latency sign differs between sessions (\(-2.2\%\), \(+7.2\%\)),
which also differ in driver and environment.  The kernel-only curve peaks at 8K; its lower values beyond depend on whether
warm-up calls enter the median, and the traced session recorded no clock.  In call latency, which is what a caller
observes, 2K is negative in both sessions.

\paragraph{dtype \(\times\) mask.}
At causal 8K the gain is \(+5.4\%\) at BF16, \(+5.3\%\) at FP16 and \(+12.4\%\) at FP8 (Table~\ref{tab:appx-b200}a).  Equal
tensor-core rates give similar gains and doubling the rate roughly doubles the gain, which an FP8-specific explanation
would not predict.  The BF16 gains at 4K--16K falsify the pre-registered prediction that they stay below 3\%.  Without the causal
mask the gain roughly doubles again (\(+25.8\%\) at FP8, Table~\ref{tab:appx-b200}c), consistent with softmax work no longer
being halved, although the mask also changes matrix-multiply volume, scheduling and boundary handling.  The fourth session
raised FP16 causal and FP16 and BF16 non-causal to three rounds (FP16 causal \(+5.2\%\to+5.3\%\), BF16 non-causal
\(+12.1\%\to+13.0\%\)).  Its clock varied more (1485--1965 MHz at 258--546 W).  Alternating the kernels limits, but does not
exclude, clock variation as a source of unequal effects, and it most likely explains the BF16 non-causal change.  The stock kernel takes almost the
same time at FP8 as at BF16 at 8K despite the doubled tensor-core rate, as expected if the exponential is exposed at FP8; without
counters this is an inference.

\paragraph{FA4's own parameter space.}
The sweep varies the stock mix of hardware and polynomial exponentials at FP8 causal 8K (Table~\ref{tab:appx-b200}d);
\(\beta\) is the fraction of elements leaving the hardware exponential under FA4's predicate (hardware iff
\texttt{k \% freq < freq - res}, \texttt{j >= frg\_cnt-1} or \texttt{j < start\_frg}; \texttt{res} \(=4\), the
tuning-table default; \texttt{frg\_cnt} \(=4\) at head dimension 128).  Six settings lie within \(-1.2\%\) to \(+2.7\%\) of
the default \texttt{freq 8/start 1}; both settings at \(\beta=75\%\) have speed-ups of \(-23.7\%\) and \(-24.5\%\), so the slowdown
follows the emulated fraction, not \texttt{freq} or first-fragment emulation (\texttt{start\_frg} \(=0\) at \(\beta=37.5\%\) is
the fastest setting).  Rowmax-H15 reaches \(+12.3\%\) in this sweep (a separate run, not averaged with the three-round
\(+12.4\%\)) and \(+9.3\%\) against the fastest stock setting measured, which need not be the best possible.  That setting,
\texttt{freq 2/start 0} and \texttt{freq 2/start 1} come from the fourth session and are compared with the first session's
default, so the near-default differences and the \(+9.3\%\) cross instances (all nine runs at 1965 MHz); the \(+12.3\%\) is
within one session.  Register spilling, FA4's own reason for partial emulation, would explain the slowdown but is
unconfirmed (no counters; no \texttt{ptxas} register summary from the CuTe-DSL toolchain).  That the 12.5\% setting beats the
stock 25\% runs opposite to FA4's source comments; we record it for our configuration only (torch 2.14, CUDA 13, batch 1,
16 heads).

\paragraph{Robustness, energy and external references.}
Across head counts, batch 4 and grouped-query attention (32 query and 8 key/value heads; rounds \(+12.4\%\), \(+12.5\%\),
\(+12.4\%\)) the FP8 8K gain stays between \(+11.7\%\) and \(+12.6\%\) (Table~\ref{tab:appx-b200}e).  At head dimension 64 it is
only \(+2.0\%\) to \(+3.7\%\), because FA4's tuning table has no entry there and stock FA4 falls back to a default
emulation setting, so the comparison is not like-for-like.  Board energy per forward falls by \(8.4\%\) at essentially
unchanged mean power, matching the shorter loop time (Table~\ref{tab:appx-b200}f): the saving comes from the shorter runtime
of the approximate operator, not from low-power arithmetic.  As context only, Rowmax-H15 on the BF16 causal path is faster
than cuDNN at 8K and 16K but slower at 4K (\(-5.1\%\)), as is stock FA4, and SDPA is far slower (Table~\ref{tab:appx-b200}g);
cuDNN computes softmax attention exactly with its own tiling, so this is not a controlled comparison.

\begingroup
\refstepcounter{table}%
\par\medskip\noindent\phantomsection\label{tab:appx-b200}{\small\textbf{Table~\thetable.} B200 attention-forward call latency, A0 (stock FA4) versus A1 (Rowmax-H15 primitive); 16 heads, head dimension 128, batch 1 unless stated.  Speed-up is \(100\,(\tilde t_{A0}/\tilde t_{A1}-1)\) with \(\tilde t\) the median over a kernel's pooled windows (21 per kernel in three-round cells); sd is the population standard deviation of the per-round speed-ups, a spread, not a confidence interval.  Per-cell protocols, including the non-alternating FP8 non-causal 8K cell, are in Appendix~\ref{app:benchmark}.}\par\smallskip
\addcontentsline{lot}{table}{\protect\numberline{\thetable}B200 FlashAttention-4 attention-forward call latency: A0 (stock FA4) vs A1 (FA4 with the exp2 primitive replaced by Rowmax-H15)}
\par\smallskip\noindent\textbf{(a) Main matrix, causal}\par\nopagebreak
{\scriptsize
\setlength{\tabcolsep}{2pt}
\begin{longtable}{@{}lllllll@{}}
\toprule
\textbf{dtype} & \textbf{seq} & \textbf{A0 ms (pooled median)} & \textbf{A1 ms} & \textbf{speed-up (pooled)} & \textbf{per-round sd} & \textbf{rounds} \\
\midrule
\endhead
FP8 E4M3 & 4K & 0.0621 & 0.0580 & +7.1\% & 2.13 & 3 \\
FP8 E4M3 & 8K & 0.2063 & 0.1835 & +12.4\% & 0.02 & 3 \\
FP8 E4M3 & 16K & 0.7320 & 0.6509 & +12.5\% & 0.08 & 3 \\
BF16 & 4K & 0.0586 & 0.0564 & +3.8\% & 0.93 & 3 \\
BF16 & 8K & 0.2020 & 0.1917 & +5.4\% & 0.17 & 3 \\
BF16 & 16K & 0.7871 & 0.7350 & +7.1\% & 0.52 & 3 \\
FP16 & 8K & 0.2156 & 0.2048 & +5.3\% & 0.25 & 3 \\
FP16 & 16K & 0.8039 & 0.7545 & +6.5\% & --- & 1 \\
\bottomrule
\end{longtable}
\addtocounter{table}{-1}
}
\par\smallskip\noindent\textbf{(b) FP8 causal sequence-length sweep (three rounds, alternating kernel order; 1K/2K retest: one pair, 9 windows per kernel)}\par\nopagebreak
{\scriptsize
\setlength{\tabcolsep}{2pt}
\begin{longtable}{@{}lllll@{}}
\toprule
\textbf{seq} & \textbf{pooled speed-up} & \textbf{per-round min \ldots{} max} & \textbf{sd} & \textbf{1K/2K retest (n = 9)} \\
\midrule
\endhead
1K & -3.6\% & -5.5\% \ldots{} +1.1\% & 2.97 & -2.2\% \\
2K & -2.6\% & -6.0\% \ldots{} +4.5\% & 4.53 & -3.7\% \\
4K & +7.9\% & +7.7\% \ldots{} +8.3\% & 0.25 & --- \\
8K & +12.1\% & +12.0\% \ldots{} +12.1\% & 0.06 & --- \\
16K & +12.5\% & +12.1\% \ldots{} +12.7\% & 0.28 & --- \\
32K & +9.9\% & +9.9\% \ldots{} +10.1\% & 0.08 & --- \\
\bottomrule
\end{longtable}
\addtocounter{table}{-1}
}
\par\noindent{\footnotesize Recorded SM clock: 1965 MHz for both kernels at \(\leq\)8K; at 16K A1's median was 1950 MHz; at 32K A1 ran at a median of 1845 MHz (range 1822--1867) against 1905 MHz (1867--1912) for A0.}\par
\par\smallskip\noindent\textbf{(c) Non-causal cells (the causal 8K cells are in (a))}\par\nopagebreak
{\scriptsize
\setlength{\tabcolsep}{2pt}
\begin{longtable}{@{}lllll@{}}
\toprule
\textbf{dtype} & \textbf{mask} & \textbf{seq} & \textbf{speed-up (pooled)} & \textbf{rounds} \\
\midrule
\endhead
FP8 & non-causal & 8K & +25.8\% (sd 0.03) & 3 \\
BF16 & non-causal & 8K & +13.0\% (sd 0.50) & 3 \\
FP16 & non-causal & 8K & +11.8\% (sd 0.70) & 3 \\
FP8 & non-causal & 4K & +27.2\% & 1 \\
FP8 & non-causal & 16K & +25.4\% & 1 \\
BF16 & non-causal & 4K & +11.4\% & 1 \\
BF16 & non-causal & 16K & +11.5\% & 1 \\
\bottomrule
\end{longtable}
\addtocounter{table}{-1}
}
\par\smallskip\noindent\textbf{(d) Stock-FA4 polynomial-emulation-fraction sweep vs Rowmax-H15 (FP8 causal 8K; speed-up relative to the stock configuration freq 8/start 1)}\par\nopagebreak
{\scriptsize
\setlength{\tabcolsep}{2pt}
\begin{longtable}{@{}lll@{}}
\toprule
\textbf{A0 configuration} & \textbf{emulation fraction $\beta$} & \textbf{vs stock} \\
\midrule
\endhead
freq 0 (hardware MUFU only) & 0\% & +1.4\% \\
freq 16, start 1 & 12.5\% & +1.6\% \\
freq 8, start 1 (stock) & 25\% & +0.0\% \\
freq 8, start 0 & 37.5\% & +2.7\% \\
freq 2, start 1 & 50\% & +0.2\% \\
freq 4, start 1 & 50\% & -1.2\% \\
freq 2, start 0 & 75\% & -23.7\% \\
freq 4, start 0 & 75\% & -24.5\% \\
\textbf{A1 Rowmax-H15 (full replacement)} & not applicable: the primitive is replaced, not emulated & \textbf{+12.3\%} \\
\bottomrule
\end{longtable}
\addtocounter{table}{-1}
}
\par\smallskip\noindent\textbf{(e) Shape robustness (FP8 causal unless stated)}\par\nopagebreak
{\scriptsize
\setlength{\tabcolsep}{2pt}
\begin{longtable}{@{}ll@{}}
\toprule
\textbf{configuration} & \textbf{speed-up} \\
\midrule
\endhead
8K, heads 4 & +12.6\% \\
8K, heads 8 & +12.3\% \\
8K, heads 32 & +12.1\% \\
8K, batch 4 & +11.7\% \\
8K, GQA (32 query / 8 KV heads): 0.38885 ms $\rightarrow$ 0.34583 ms & +12.4\% (per-round sd 0.03) \\
head dim 64, FP8: 4K / 8K / 16K & +2.0\% / +2.5\% / +3.7\% \\
head dim 64, BF16: 4K / 8K / 16K & +0.7\% / +7.6\% / +9.7\% \\
2K / 32K (extension run) & +7.8\% / +10.4\% \\
\bottomrule
\end{longtable}
\addtocounter{table}{-1}
}
\par\noindent{\footnotesize The stock FA4 tuning table has no head-dim-64 entry, so A0 falls back to a default emulation setting there; the head-dim-64 cells are therefore not a like-for-like comparison with head dim 128.}\par
\par\smallskip\noindent\textbf{(f) Sustained power and energy (FP8 causal 16K, 45 s sustained forward loop)}\par\nopagebreak
{\scriptsize
\setlength{\tabcolsep}{2pt}
\begin{longtable}{@{}lllll@{}}
\toprule
\textbf{kernel} & \textbf{forwards} & \textbf{mean power (W)} & \textbf{time per forward (ms)} & \textbf{energy per forward (mJ)} \\
\midrule
\endhead
A0 stock FA4 & 59550 & 982.3 & 0.7559 & 742.5 \\
A1 FA4 + Rowmax-H15 & 65050 & 982.9 & 0.6922 & 680.3 \\
\bottomrule
\end{longtable}
\addtocounter{table}{-1}
}
\par\noindent{\footnotesize Energy and time per forward both change by -8.4\% at unchanged mean power: the saving follows from the shorter runtime, not from a low-power effect.}\par
\par\smallskip\noindent\textbf{(g) External references (BF16, median ms; different implementation strategies --- context, not the controlled baseline)}\par\nopagebreak
{\scriptsize
\setlength{\tabcolsep}{2pt}
\begin{longtable}{@{}>{\raggedright\arraybackslash}p{\dimexpr 0.082\linewidth-4pt\relax}>{\raggedright\arraybackslash}p{\dimexpr 0.137\linewidth-4pt\relax}>{\raggedright\arraybackslash}p{\dimexpr 0.186\linewidth-4pt\relax}>{\raggedright\arraybackslash}p{\dimexpr 0.096\linewidth-4pt\relax}>{\raggedright\arraybackslash}p{\dimexpr 0.124\linewidth-4pt\relax}>{\raggedright\arraybackslash}p{\dimexpr 0.124\linewidth-4pt\relax}>{\raggedright\arraybackslash}p{\dimexpr 0.111\linewidth-4pt\relax}>{\raggedright\arraybackslash}p{\dimexpr 0.111\linewidth-4pt\relax}@{}}
\toprule
\textbf{seq} & \textbf{torch SDPA flash} & \textbf{torch SDPA mem-efficient} & \textbf{cuDNN 9.24} & \textbf{FA4 stock (A0)} & \textbf{FA4 + Rowmax-H15 (A1)} & \textbf{Rowmax-H15 vs cuDNN} & \textbf{FA4 vs cuDNN} \\
\midrule
\endhead
4K & 0.2876 & 0.5596 & 0.0536 & 0.0586 & 0.0564 & -5.1\% & -8.6\% \\
8K & 0.9117 & 1.9075 & 0.2230 & 0.2020 & 0.1917 & +16.4\% & +10.4\% \\
16K & 3.1568 & 6.9877 & 0.8316 & 0.7871 & 0.7350 & +13.1\% & +5.6\% \\
\bottomrule
\end{longtable}
\addtocounter{table}{-1}
}
\par\noindent{\footnotesize cuDNN and SDPA compute softmax attention; Rowmax-H15 is approximate.  BF16 only (no FP8 SDPA path), the dtype with the smallest Rowmax-H15 gain.}\par
\endgroup

\FloatBarrier
\subsection{Mechanism and limitations}
\label{app:sass}\label{app:predictions}\label{sec:realization-scope}

\paragraph{Static instruction audit.}
Isolated micro-kernels compiled for sm\_86 give 8 instructions per element for FA4's cubic emulation (7 on the FP
pipeline) against 3 for Rowmax-H15 (2), and dependency depth 8 against 3; the single-application kernels differ by two
registers per thread and the 32-element fragment kernels both use 40.  This concerns the operator in isolation, not the
fused B200 kernel, and does not rule out register pressure or another mechanism there.

\paragraph{Counters and pre-registered predictions.}
Counter-level profiling (\texttt{ncu}) was refused with \texttt{ERR\_NVGPUCTRPERM} on all three instances where it was
attempted (two driver versions; one with a full Nsight Compute 2025.1.1 installation), and sampled GPU metrics were
unavailable, so the mechanism is not verified by counters.  Six predictions were sealed (SHA-256) before any B200 run.
Predictions 1--3 and 6 (MUFU utilization near zero, FP-pipeline utilization up, softmax-stage time \(-35\)--\(50\%\),
registers per thread A1 \(\leq\) A0) need counters and remain unverified.  Prediction 4, an FP8 causal hd128 gain of
\(+10\)--\(15\%\) for the whole attention forward (``end-to-end'' in the sealed file), was met (\(+12.4\%\) at 8K, \(+12.5\%\) at
16K, call latency).  Prediction 5, a BF16 whole-forward gain below 3\%, failed (\(+3.8\%\), \(+5.4\%\), \(+7.1\%\) at 4K--16K).
The record redefines no endpoint.

\paragraph{Scope.}
Measured kernel fidelity covers five models on the BF16 path at 2K; the FP8 figure is a Qwen2.5-1.5B semantic simulation; no
fidelity was measured at the 8K and 16K timing shapes; and the ten-model Rowmax-PoT evidence is not a ten-model kernel
result.  Timing covers the B200 attention forward only---not backward, decode, head dimensions 192/256 or full-model
latency.  Counting decoder-layer matrix multiplies without the language-model head, the causal attention forward
(\(2s^2d\) FLOPs per layer, against \(2s(2d^2+2d\,d_{kv}+3df)\) for projections and MLP) is 21\% of Qwen2.5-1.5B's
prefill FLOPs at 8K and 35\% at 16K (\(d=1536\), 28 layers, 12 query and 2 key/value heads of width 128, FFN width 8960
recovered from the parameter count); these are workload shares, not runtime shares, and give no prefill speed-up, which
we did not measure.  FA4 \citep{flashattention4-2026} reports doubled native exponential throughput on B300/GB300 (32
ops/clk/SM), and its tuning table disables polynomial emulation for sm\_103; we make no claim about B300.

\FloatBarrier
\section{Long-context and downstream evaluation}
\label{app:breadth}\label{sec:robustness}
\setcounter{table}{0}\renewcommand{\thetable}{C\arabic{table}}\renewcommand{\theHtable}{C\arabic{table}}

\FloatBarrier
\subsection{Setup}
\label{sec:methods-breadth}\label{sec:appendix-70b-execution}

\paragraph{Long context.}
Qwen2.5-1.5B, Qwen2.5-72B and Llama-3.1-70B are evaluated at \(L\in\{2048,8192,16384\}\) and Llama-3.2-3B at
\(L\in\{2048,8192\}\); the 200,000-token stream gives 97, 24 and 12 complete blocks (198,559, 196,584 and 196,596
predictions).  Each context is a separate experiment: its softmax baseline is recomputed, its reference level \(J_0\) (PoT \(m=1\))
is remeasured over \(\max\{1,\lfloor 32768/L\rfloor\}\) divergence windows, and its flattening and sharpening
temperatures are re-solved against that level.  The conditions are softmax, \(K=21\) allocation at \(R=1\) and \(R=4\),
PoT \(m=1\) and the two matched temperatures, so \(D_R\) and \(D_T\) are paired only within a context.  The two larger models ran
on three A100-SXM4-80GB GPUs per job (Table~\ref{tab:appx-env}), an execution detail, not a performance experiment.

\paragraph{Zero-shot.}
Qwen2.5-1.5B and both larger models are evaluated with lm-evaluation-harness (0.4.12 for the larger models), zero demonstrations,
batch size eight with left padding after a batch-versus-solo attention-mask gate, and the full evaluation sets of
LAMBADA OpenAI, HellaSwag, ARC-Easy, ARC-Challenge, PIQA and WinoGrande.  The conditions are softmax, mean-threshold
support with softmax weights, full-range exponential \(K=32,R=4\) and \(K=64,R=4\), and Rowmax-PoT \(K_{\max}=20\).  One
metric set is used throughout: accuracy for LAMBADA and WinoGrande and length-normalized accuracy (\texttt{acc\_norm})
for the other four; LAMBADA perplexity is also recorded.  Comparisons are descriptive, within task and metric; harness
standard errors describe single estimates, not paired differences, and no aggregate score or paired endpoint is defined.

\paragraph{Vocabulary output.}
With softmax-baseline and condition vocabulary distributions \(b_t\), \(c_t\) at prediction \(t\) and
\(m_t=(b_t+c_t)/2\),
\begin{align}
  \operatorname{KL}_{\mathrm{fwd}}&=\mathbb E_t\!\left[\operatorname{KL}(b_t\|c_t)\right],
  &\operatorname{KL}_{\mathrm{rev}}&=\mathbb E_t\!\left[\operatorname{KL}(c_t\|b_t)\right],\\
  \operatorname{JSD}_{\mathrm{vocab}}&=\mathbb E_t\!\left[\tfrac12\operatorname{KL}(b_t\|m_t)+\tfrac12\operatorname{KL}(c_t\|m_t)\right],
  &A_{\mathrm{top1}}&=\mathbb E_t\!\left[\mathbf 1\{\arg\max b_t=\arg\max c_t\}\right],
  \label{eq:vocab-fidelity}
\end{align}
averaged within block and then across blocks with prediction-count weights.  The summaries carry no paired intervals,
and the correlations across conditions are descriptive diagnostics, not endpoints.

\FloatBarrier
\subsection{Long-context results}
\AppendixFigure{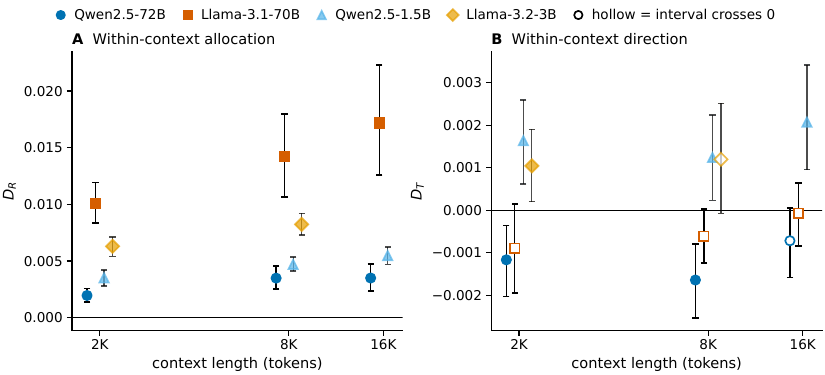}
  {Long-context within-context robustness.}
  {(A)~Allocation contrast \(D_R\) and (B)~matched-JSD contrast \(D_T\), re-matched within each context length, with 95\%
  paired block-bootstrap intervals; filled if the interval excludes zero, hollow if it includes zero.  Each context has
  its own baseline, \(J_0\) and temperatures; no cross-context comparison is made.}
  {fig:long-context}
\begin{table}[H]
  \centering
  \small
  \setlength{\tabcolsep}{4.0pt}
  \caption{\textbf{Per-context matched-JSD reproducibility for the two
  approximately 70B models.} Each context has its own softmax baseline, reference
  level $J_0$ (PoT $m=1$), and independently resolved flattening ($A^-$) and
  sharpening ($A^+$) temperatures. Parentheses give relative target error in
  percent; all twelve matches are within 1\%. Contexts are not compared by a
  cross-context significance test.}
  \label{tab:70b-t11-matching}
  \begin{tabular}{@{}llrrrrr@{}}
    \toprule
    Model & Context & Blocks & Predictions & $J_0$ &
    \shortstack{$A^-$ achieved\\(rel. err. \%)} &
    \shortstack{$A^+$ achieved\\(rel. err. \%)} \\
    \midrule
  Qwen2.5-72B & 2K & 97 & 198,559 & 0.001228 & 0.001229 (0.084) & 0.001235 (0.591) \\
  Qwen2.5-72B & 8K & 24 & 196,584 & 0.001323 & 0.001315 (0.614) & 0.001333 (0.725) \\
  Qwen2.5-72B & 16K & 12 & 196,596 & 0.001362 & 0.001371 (0.629) & 0.001353 (0.666) \\
  Llama-3.1-70B & 2K & 97 & 198,559 & 0.001778 & 0.001787 (0.510) & 0.001785 (0.393) \\
  Llama-3.1-70B & 8K & 24 & 196,584 & 0.001137 & 0.001129 (0.687) & 0.001138 (0.100) \\
  Llama-3.1-70B & 16K & 12 & 196,596 & 0.001236 & 0.001244 (0.621) & 0.001237 (0.062) \\
    \bottomrule
  \end{tabular}
\end{table}

Allocation holds within every context (Figure~\ref{fig:long-context}A).  In Qwen2.5-1.5B, \(D_R\) is \(+0.00351\)
\([+0.00282,+0.00420]\), \(+0.00470\) \([+0.00413,+0.00535]\) and \(+0.00548\) \([+0.00470,+0.00623]\) at 2K, 8K and 16K;
in Llama-3.2-3B it is \(+0.00629\) \([+0.00540,+0.00714]\) and \(+0.00823\) \([+0.00727,+0.00919]\) at 2K and 8K.  In
Qwen2.5-72B it is \(0.00196\) \([0.00137,0.00258]\), \(0.00349\) \([0.00250,0.00453]\) and \(0.00350\) \([0.00234,0.00474]\),
and in Llama-3.1-70B \(0.0101\) \([0.00833,0.0119]\), \(0.0142\) \([0.0107,0.0180]\) and \(0.0172\) \([0.0126,0.0223]\), so
all six cells of the larger models are resolved.  \(R=4\) removes 67.3\%, 78.1\% and 72.9\% (Qwen2.5-72B) and 64.9\%, 70.5\% and 69.9\% (Llama-3.1-70B)
of the \(R=1\) degradation; at 16K it also moves output JSD from 0.001538 to 0.000627 and top-1 agreement from 97.15\% to
98.11\% in Qwen2.5-72B, and from 0.006584 to 0.002269 and 95.74\% to 97.44\% in Llama-3.1-70B (descriptive).

The directional contrast is weaker and model-dependent (Figure~\ref{fig:long-context}B).  Among the cells of the larger models it is
resolved only for Qwen2.5-72B, negative, at 2K (\(-0.00116\) \([-0.00203,-0.000357]\)) and 8K (\(-0.00164\)
\([-0.00252,-0.000798]\)); its 16K interval crosses zero (\(-0.000716\) \([-0.00158,+0.0000558]\)), and all three
Llama-3.1-70B estimates are negative but unresolved: \(-0.000888\) \([-0.00195,+0.000142]\), \(-0.000611\)
\([-0.00123,+0.0000341]\) and \(-0.0000675\) \([-0.000841,+0.000646]\).  Qwen2.5-1.5B stays positive and resolved
(\(+0.00164\), \(+0.00124\), \(+0.00208\)), and Llama-3.2-3B's positive 8K estimate is unresolved.  Long-context
transfer is therefore clear for allocation but not for a stable directional effect.  All twelve matches of the larger models are within
1\% relative target error (maximum 0.725\%; Table~\ref{tab:70b-t11-matching}).  These are within-context effects: they do
not test change across context lengths and are not pooled across models or contexts.

\FloatBarrier
\subsection{Zero-shot results}
\begin{table}[!htbp]
  \centering
  \scriptsize
  \setlength{\tabcolsep}{2pt}
  \caption{\textbf{Zero-shot scores under attention approximations.}  Each cell is the score in percent followed by
  its descriptive change from the same-model softmax baseline in percentage points.  Accuracy for LAMBADA and
  WinoGrande; length-normalized accuracy (\texttt{acc\_norm}) for HellaSwag, ARC-Easy, ARC-Challenge and PIQA.
  The last column is the LAMBADA perplexity change (\%) relative to softmax, reported for the two larger models.  No pooled score or paired
  cross-condition interval is defined.}
  \label{tab:appx-zeroshot}\label{tab:zero-shot-breadth}\label{tab:70b-zero-shot-full}\label{tab:70b-zero-shot-summary}
  \begin{tabular}{@{}lrrrrrrr@{}}
    \toprule
    Condition & LAMBADA & HellaSwag & ARC-E & ARC-C & PIQA & WinoGrande & \shortstack{LAMBADA\\PPL (\%)} \\
    \midrule
    \multicolumn{8}{@{}l}{\textbf{Qwen2.5-72B}} \\
    \quad Softmax & 77.72 (+0.00) & 86.10 (+0.00) & 83.38 (+0.00) & 62.71 (+0.00) & 83.62 (+0.00) & 77.35 (+0.00) & --- \\
    \quad Mean-threshold support & 76.73 ($-$0.99) & 85.72 ($-$0.38) & 84.81 (+1.43) & 62.71 (+0.00) & 81.83 ($-$1.80) & 76.16 ($-$1.18) & +6.727 \\
    \quad Exp.\ $K=32,R=4$ & 77.59 ($-$0.14) & 86.13 (+0.03) & 83.25 ($-$0.13) & 62.54 ($-$0.17) & 83.62 (+0.00) & 77.35 (+0.00) & +0.037 \\
    \quad Exp.\ $K=64,R=4$ & 77.64 ($-$0.08) & 86.05 ($-$0.05) & 83.21 ($-$0.17) & 62.63 ($-$0.09) & 83.68 (+0.05) & 78.30 (+0.95) & +0.033 \\
    \quad Rowmax-PoT $K_{\max}=20$ & 77.53 ($-$0.19) & 86.06 ($-$0.04) & 83.46 (+0.08) & 62.20 ($-$0.51) & 83.73 (+0.11) & 77.74 (+0.39) & +0.231 \\
    \addlinespace
    \multicolumn{8}{@{}l}{\textbf{Llama-3.1-70B}} \\
    \quad Softmax & 79.53 (+0.00) & 85.75 (+0.00) & 81.44 (+0.00) & 61.52 (+0.00) & 84.28 (+0.00) & 81.69 (+0.00) & --- \\
    \quad Mean-threshold support & 77.29 ($-$2.23) & 82.64 ($-$3.11) & 82.53 (+1.09) & 59.81 ($-$1.71) & 83.30 ($-$0.98) & 77.51 ($-$4.18) & +11.649 \\
    \quad Exp.\ $K=32,R=4$ & 79.39 ($-$0.14) & 85.63 ($-$0.12) & 81.57 (+0.13) & 61.26 ($-$0.26) & 84.17 ($-$0.11) & 81.85 (+0.16) & +0.143 \\
    \quad Exp.\ $K=64,R=4$ & 79.35 ($-$0.17) & 85.74 ($-$0.01) & 81.73 (+0.29) & 61.18 ($-$0.34) & 84.33 (+0.05) & 82.00 (+0.32) & +0.101 \\
    \quad Rowmax-PoT $K_{\max}=20$ & 79.37 ($-$0.16) & 85.44 ($-$0.31) & 81.14 ($-$0.29) & 61.43 ($-$0.09) & 84.44 (+0.16) & 81.14 ($-$0.55) & +0.855 \\
    \addlinespace
    \multicolumn{8}{@{}l}{\textbf{Qwen2.5-1.5B}} \\
    \quad Softmax & 62.22 (+0.00) & 67.90 (+0.00) & 71.80 (+0.00) & 45.22 (+0.00) & 75.95 (+0.00) & 63.77 (+0.00) & --- \\
    \quad Mean-threshold support & 62.16 ($-$0.06) & 67.36 ($-$0.55) & 73.78 (+1.98) & 44.97 ($-$0.26) & 76.06 (+0.11) & 62.51 ($-$1.26) & --- \\
    \quad Exp.\ $K=32,R=4$ & 62.24 (+0.02) & 67.75 ($-$0.16) & 71.84 (+0.04) & 45.39 (+0.17) & 75.79 ($-$0.16) & 64.09 (+0.32) & --- \\
    \quad Exp.\ $K=64,R=4$ & 62.27 (+0.06) & 67.82 ($-$0.09) & 72.10 (+0.29) & 44.37 ($-$0.85) & 75.95 (+0.00) & 63.77 (+0.00) & --- \\
    \quad Rowmax-PoT $K_{\max}=20$ & 61.83 ($-$0.39) & 67.80 ($-$0.11) & 71.76 ($-$0.04) & 44.88 ($-$0.34) & 75.73 ($-$0.22) & 63.69 ($-$0.08) & --- \\
    \bottomrule
  \end{tabular}
\end{table}

For the two larger models the \(K=32,R=4\) condition changes the six scores by a mean absolute 0.114 percentage points and at
most 0.256 over the 12 model--task cells (ARC-Challenge \(-0.171\) and \(-0.256\)), while LAMBADA perplexity changes by
\(+0.037\%\) and \(+0.143\%\) (Table~\ref{tab:appx-zeroshot}); Qwen2.5-1.5B's mean absolute change is 0.15 points.
The \(K=64,R=4\) and Rowmax-PoT conditions also change scores modestly but are not uniformly closer to baseline
(Rowmax-PoT raises LAMBADA perplexity by \(+0.231\%\) and \(+0.855\%\)).  Mean-threshold support is the negative control: LAMBADA perplexity rises by \(6.727\%\)
and \(11.649\%\), and Llama-3.1-70B loses several points on some tasks (HellaSwag \(-3.11\), maximum absolute change
4.18).  Taskwise directions are mixed; the check shows that this \(K=32,R=4\) condition preserved these six metrics
closely, not broad capability preservation or benchmark superiority.

\FloatBarrier
\subsection{Vocabulary-output fidelity}

Over the 137 completed default-context conditions of Qwen2.5-1.5B, vocabulary-output JSD tracks
\(\Delta\mathrm{NLL}\) with Spearman \(\rho=+0.994\) (Pearson \(+0.995\)) and top-1 agreement with \(\rho=-0.991\); over the
ten models the JSD Spearman correlation is at least 0.546 and the top-1 correlation at most \(-0.544\).  At the common
\(K=32,R=4\) point the ten models span forward KL \(0.000778\)--\(0.00887\), reverse KL \(0.000780\)--\(0.00880\), JSD
\(0.000194\)--\(0.00215\) and top-1 agreement \(95.8\%\)--\(98.7\%\); at Rowmax-PoT \(K_{\max}=20\), \(0.00329\)--\(0.0284\),
\(0.00336\)--\(0.0283\), \(0.000808\)--\(0.00681\) and \(92.1\%\)--\(97.6\%\).  Gemma-2-2B is the least faithful in both, with
the \(92.1\%\) minimum top-1 agreement at only a \(0.353\%\) perplexity increase, so a small average NLL change does not
imply identical predictive distributions.  Over the 18 condition summaries of each larger model, \(\Delta\mathrm{NLL}\)
correlates with forward KL, reverse KL, JSD and top-1 agreement at Pearson \(0.967\), \(0.972\), \(0.971\) and \(-0.921\)
(Qwen2.5-72B) and \(0.999\), \(0.998\), \(0.999\) and \(-0.972\) (Llama-3.1-70B).  These are within-model, in-sample
associations without intervals, not held-out predictors or evidence of output equivalence; attention-distribution JSD
(Appendix~\ref{sec:results-structure}) and vocabulary-output JSD are distinct quantities.

\FloatBarrier
\section{Additional characterization and mechanism evidence}
\label{app:structure}\label{sec:results}\label{app:char-tables}
\setcounter{table}{0}\renewcommand{\thetable}{D\arabic{table}}\renewcommand{\theHtable}{D\arabic{table}}

\FloatBarrier
\subsection{Amount}
\label{sec:results-amount}\label{app:support}
\AppendixFigure{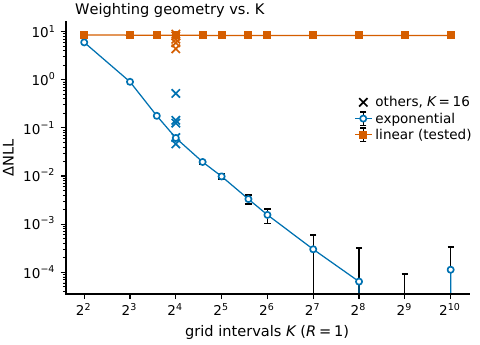}
  {Weighting geometry under full support.}
  {\(\Delta\mathrm{NLL}\) of the exponential and the tested linear level-to-weight maps against \(K\) (full support,
  \(R=1\), upper-edge reconstruction): Qwen2.5-1.5B (lines) and the other models with this task at \(K=16\) (crosses);
  95\% paired-bootstrap intervals.}
  {fig:amount-geometry-supp}

\paragraph{Support.}
Under softmax weights the Qwen2.5-1.5B top-\(k\) sweep raises perplexity by \(0.131\%\) (50\% of valid entries kept;
\(\Delta\mathrm{NLL}=0.00130\)), \(0.828\%\) (25\%; \(0.00825\)), \(3.092\%\) (15\%; \(0.0304\)), \(7.821\%\) (10\%;
\(0.0753\)), \(15.185\%\) (6.5\%), \(37.861\%\) (2.6\%), \(90.478\%\) (1\%) and \(144.670\%\) (0.65\%); the last two
targets retain pooled densities of \(1.05\%\) and \(0.70\%\).  The mean-threshold rule keeps a fraction 0.493 and costs
\(\Delta\mathrm{NLL}=0.00309\) (\(0.310\%\)).  Uniform weights with the same support rules multiply
perplexity by \(5268\) when retaining half of the valid positions and by \(4155\) at the mean threshold.  Because these are all-layer interventions, the
strict single-layer control isolates the weighting: \(D_W(\ell)\) ranges from \(0.0181\) to \(0.2000\) over layers 1--27 with
layer~0 at \(8.1733\), and the block-paired aggregate over the fixed layer set is \(0.344\) \([0.340,0.348]\).  The
layerwise result (28 of 28 positive, 28 of 28 pointwise intervals above zero) is the primary summary rather than this
layer-0-dominated aggregate.

\paragraph{Weighting geometry and precision.}
Under full support with \(R=1\) and upper-edge reconstruction, the exponential grid's \(\Delta\mathrm{NLL}\) falls from
\(0.0617\) at \(K=16\) to \(0.00981\) at \(K=32\) and \(0.00156\) at \(K=64\), while the tested linear map stays between
\(8.32\) and \(8.37\) (Figure~\ref{fig:amount-geometry-supp}); the paired linear-minus-exponential contrast at \(K=32\) is
\(8.33\) \([8.15,8.51]\), and in the six models with this task the linear-grid damage at \(K=16\) lies between \(4.39\) and
\(8.86\).  This constrains the tested families; it does not make exponential weighting uniquely optimal.  With nearest
reconstruction and \(R=4\), Qwen2.5-1.5B costs \(0.281\%\), \(0.091\%\) and \(0.00733\%\) perplexity at \(K=16\),
\(32\) and \(64\); the \(K=32\) point is then replicated descriptively across models (Table~\ref{tab:crossmodel-summary}).

\FloatBarrier
\subsection{Allocation}
\label{sec:results-allocation}

Attention mass is highly concentrated on Qwen2.5-1.5B's softmax trajectory: entries with \(p\geq0.1\) are
\(0.151\%\) of valid entries and carry \(53.8\%\) of the mass, entries with \(p\geq0.01\) are \(0.985\%\) and carry
\(76.9\%\), entries with \(p\geq0.001\) are \(6.34\%\) and carry \(92.5\%\), and entries with \(p<0.0001\) are \(75.1\%\)
and carry \(1.2\%\).  Across \(R\in\{0.25,0.5,1,2,4,8\}\) Qwen2.5-1.5B's
\(\Delta\mathrm{NLL}\) falls monotonically, \(0.0161\), \(0.00997\), \(0.00489\), \(0.00251\), \(0.00138\), \(0.00100\), so the
anti-allocation points are worse than uniform and the tested \(R>1\) points better.  Along the same sweep the unweighted
RMS log-weight error rises from \(0.314\) (\(R=1\)) to \(0.401\) (\(R=4\)) and \(0.483\) (\(R=8\)) while the \(p\)-weighted
error falls from \(0.178\) to \(0.126\) and \(0.124\); relative to \(R=1\), the ratios are \(1.279\), \(0.708\) and \(0.283\)
(for \(\Delta\mathrm{NLL}\)) at \(R=4\) and \(1.539\), \(0.698\) and \(0.205\) at \(R=8\).  The error summaries are measured on
the softmax trajectory, whereas \(\Delta\mathrm{NLL}\) comes from applying the approximation throughout the network, so a
representation can be less accurate over all entries yet better for the model.  Table~\ref{tab:appx-dr} lists
\(D_R(21)\) and \(I_R\) per model.

\begingroup
\refstepcounter{table}%
\par\medskip\noindent\phantomsection\label{tab:appx-dr}{\small\textbf{Table~\thetable.} Allocation contrast $D_R$(21) = $\Delta$NLL(R=1) $-$ $\Delta$NLL(R=4) at K=21 (paired block bootstrap, 5000 replicates, seed 0) and the precision $\times$ allocation interaction $I_R$ = $D_R$(16) $-$ $D_R$(32) where K=16/32 cells exist. Models are listed individually; no pooled cross-model effect is defined.}\par\smallskip
\addcontentsline{lot}{table}{\protect\numberline{\thetable}Allocation contrast $D_R$(21) = $\Delta$NLL(R=1) $-$ $\Delta$NLL(R=4) at K=21 (paired block bootstrap, 5000 replicates, seed 0) and the precision $\times$ allocation interaction $I_R$ = $D_R$(16) $-$ $D_R$(32) where K=16/32 cells exist}
{\scriptsize
\setlength{\tabcolsep}{2pt}
\begin{longtable}{@{}lll>{\raggedright\arraybackslash}p{0.08\linewidth}lll@{}}
\toprule
\textbf{model} & \textbf{params (B)} & \textbf{$D_R$(21) [95\% CI]} & \textbf{resolved?} & \textbf{$D_R$(16)} & \textbf{$D_R$(32)} & \textbf{$I_R$ [95\% CI]} \\
\midrule
\endhead
Qwen2.5-0.5B & 0.49 & +0.00611 [+0.00525, +0.00697] & yes & +0.0105 & +0.00192 & +0.00855 [+0.00712, +0.0101] \\
Llama-3.2-1B & 1.24 & +0.0109 [+0.00976, +0.0120] & yes & --- & --- & --- \\
Qwen2.5-1.5B & 1.54 & +0.00351 [+0.00282, +0.00420] & yes & +0.00612 & +0.00133 & +0.00480 [+0.00356, +0.00600] \\
Gemma-2-2B & 2.61 & +0.00527 [-0.00369, +0.0132] & \textbf{no (interval crosses 0)} & --- & --- & --- \\
Qwen2.5-3B & 3.09 & +0.00331 [+0.00256, +0.00404] & yes & +0.00532 & +0.00113 & +0.00419 [+0.00305, +0.00535] \\
Llama-3.2-3B & 3.21 & +0.00629 [+0.00540, +0.00714] & yes & --- & --- & --- \\
Mistral-7B-v0.3 & 7.25 & +0.00167 [+0.00119, +0.00218] & yes & --- & --- & --- \\
Llama-3.1-8B & 8.03 & +0.00462 [+0.00383, +0.00542] & yes & --- & --- & --- \\
Llama-3.1-70B & 70.55 & +0.0101 [+0.00833, +0.0119] & yes & +0.0176 & +0.00343 & +0.0142 [+0.0113, +0.0171] \\
Qwen2.5-72B & 72.71 & +0.00196 [+0.00137, +0.00258] & yes & +0.00380 & +0.000788 & +0.00302 [+0.00200, +0.00409] \\
\bottomrule
\end{longtable}
\addtocounter{table}{-1}
}
\par\noindent{\footnotesize Point estimates positive in 10/10 models; intervals exclude zero in 9/10 (unresolved: Gemma-2-2B). The interaction is resolved (interval above zero) in 4/4 Qwen2.5 scales; Llama-3.1-70B provides a separate model-specific corroboration.}\par
\endgroup

\FloatBarrier
\subsection{Structure}
\label{sec:results-structure}

The Qwen2.5-1.5B response map has 56 interventions (full-range exponential quantization and allocation,
power-of-two perturbations, temperature flattening and sharpening), all with positive mean row-wise JSD \(\overline J\)
and positive \(\Delta\mathrm{NLL}\); the in-sample fit is
\(\log\Delta\mathrm{NLL}=0.465+1.050\log\overline J\) (\(R^2=0.945\), \(n=56\)).  Table~\ref{tab:appx-dt} gives the
matched-JSD and matched-TV contrasts per model, including the resolved negative \(D_T(J_0)\) of Llama-3.2-1B and
Qwen2.5-72B and the unresolved Llama-3.1-70B.

\begingroup
\refstepcounter{table}%
\par\medskip\noindent\phantomsection\label{tab:appx-dt}{\small\textbf{Table~\thetable.} Matched-distortion directional contrasts $D_T$ = $\Delta$NLL(A$^-$) $-$ $\Delta$NLL(A$^+$) with flattening (A$^-$) and sharpening (A$^+$) independently matched to the mean row-wise JSD J$_0$ of the octave PoT perturbation and to 10J$_0$; the matched-TV robustness check $D_{TV}$ for the two selected models is given below the table. Paired block bootstrap; 'resolved' = interval excludes zero.}\par\smallskip
\addcontentsline{lot}{table}{\protect\numberline{\thetable}Matched-distortion directional contrasts $D_T$ = $\Delta$NLL(A$^-$) $-$ $\Delta$NLL(A$^+$) with flattening (A$^-$) and sharpening (A$^+$) independently matched to the mean row-wise JSD J$_0$ of the octave PoT perturbation and to 10J$_0$; the matched-TV robustness check $D_{TV}$ for the two selected models is given below the table}
{\scriptsize
\renewcommand{\theHtable}{\thetable.lt\arabic{LT@tables}}%
\setlength{\tabcolsep}{4pt}
\begin{longtable}{@{}lllll@{}}
\toprule
\textbf{model} & \textbf{$D_T$(J$_0$) [95\% CI]} & \textbf{resolved} & \textbf{$D_T$(10J$_0$) [95\% CI]} & \textbf{resolved} \\
\midrule
\endhead
Qwen2.5-0.5B & +0.00472 [+0.00331, +0.00607] & + & +0.0300 [+0.0258, +0.0341] & + \\
Llama-3.2-1B & -0.00313 [-0.00428, -0.00202] & $-$ (resolved negative) & -0.000204 [-0.00339, +0.00285] & unresolved \\
Qwen2.5-1.5B & +0.00170 [+0.000676, +0.00269] & + & +0.0133 [+0.0104, +0.0161] & + \\
Gemma-2-2B & +0.246 [+0.218, +0.273] & + & +0.806 [+0.767, +0.844] & + \\
Qwen2.5-3B & +0.00409 [+0.00308, +0.00512] & + & +0.0195 [+0.0164, +0.0227] & + \\
Llama-3.2-3B & +0.000976 [+0.000138, +0.00185] & + & +0.0107 [+0.00813, +0.0135] & + \\
Mistral-7B-v0.3 & +0.00266 [+0.00192, +0.00336] & + & +0.0183 [+0.0158, +0.0207] & + \\
Llama-3.1-8B & +0.000802 [+0.0000767, +0.00154] & + & +0.00732 [+0.00509, +0.00953] & + \\
Llama-3.1-70B & -0.00107 [-0.00220, +0.0000282] & unresolved & +0.0281 [+0.0232, +0.0331] & + \\
Qwen2.5-72B & -0.00119 [-0.00208, -0.000375] & $-$ (resolved negative) & +0.000628 [-0.00190, +0.00302] & unresolved \\
\bottomrule
\end{longtable}
\addtocounter{table}{-1}
}
\par\noindent{\footnotesize Matched TV, $D_{TV}$(J$_0$) and $D_{TV}$(10J$_0$) [95\% CI]: Llama-3.2-1B -0.00282 [-0.00382, -0.00182] and -0.000261 [-0.00329, +0.00267]; Qwen2.5-1.5B +0.00137 [+0.000375, +0.00232] and +0.0111 [+0.00827, +0.0137].}\par
\par\noindent{\footnotesize Resolved negative J$_0$ contrasts (sharpening more damaging than flattening at matched JSD): Qwen2.5-72B, Llama-3.2-1B. Unresolved at J$_0$: Llama-3.1-70B. Unresolved at 10J$_0$: Qwen2.5-72B, Llama-3.2-1B. Qwen2.5-1.5B 3J$_0$ contrast: +0.00352 [+0.00184, +0.00510].}\par
\endgroup

\FloatBarrier
\subsection{Redistribution geometry and rescue}
\label{sec:results-geometry}
\AppendixFigure{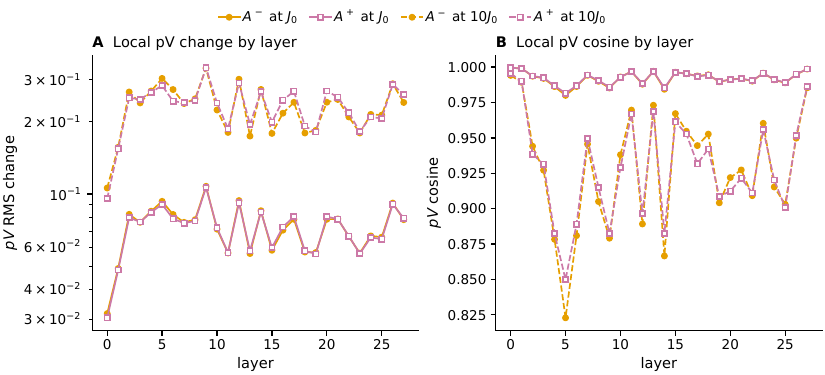}
  {Local attention-output magnitude as a negative control.}
  {Per-layer (A)~RMS change and (B)~cosine of the local output \(pV\) for matched flattening (\(A^-\)) and sharpening
  (\(A^+\)) at \(J_0\) (solid) and \(10J_0\) (dashed), Qwen2.5-1.5B; descriptive aggregates without bootstrap intervals.}
  {fig:local-pv-supp}
\AppendixFigure{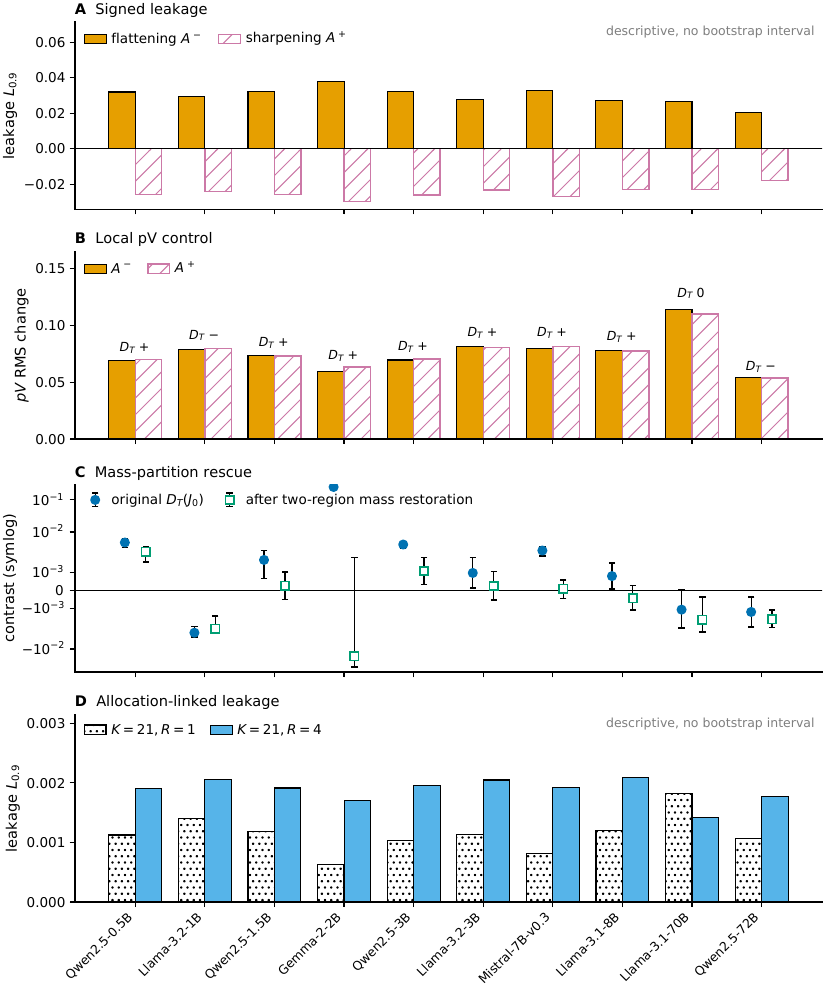}
  {Similar probability redistribution, model-dependent losses.}
  {(A)~Signed leakage \(L_{0.9}\) of matched flattening (filled) and sharpening (hatched) at \(J_0\).  (B)~Local \(pV\) RMS
  change for the same conditions, with the sign of the resolved \(D_T\) above each pair.  (C)~Original \(D_T(J_0)\)
  (circles) and the contrast after two-region mass restoration (hollow squares), 95\% paired intervals, symlog axis;
  \(G_{\mathrm{gap}}\) is their signed difference.  (D)~\(L_{0.9}\) at \(K=21\) for \(R=1\) (dotted) and \(R=4\) (filled).
  Panels A, B and D are descriptive, without intervals.}
  {fig:geometry}

\paragraph{Shared redistribution geometry.}
In all ten models matched flattening moves probability out of \(S_{0.9}\) and matched sharpening moves it in
(Figure~\ref{fig:geometry}A): at \(J_0\), flattening leakage ranges from \(0.02022\) to \(0.03768\) and sharpening leakage
from \(-0.02965\) to \(-0.01792\), and at \(10J_0\) over \([0.07187,0.14562]\) and \([-0.07140,-0.04883]\).  Llama-3.2-1B
shows the same pattern (\(+0.0294\) for \(A^-\), \(-0.0242\) for \(A^+\)) despite its opposite downstream contrast, so the
same redistribution can come with a different loss response.  These summaries cover four blocks and have no intervals.

\paragraph{Local \(pV\) is a negative control.}
The local output RMS change is nearly identical for the two directions (Qwen2.5-1.5B 0.0733 against 0.0732;
Llama-3.2-1B 0.0790 against 0.0798; Figures~\ref{fig:geometry}B and~\ref{fig:local-pv-supp}).  At \(J_0\) the median
absolute difference between the two directions is \(0.94\%\) of the smaller value (maximum \(7.17\%\)), and the direction with the larger RMS is sharpening in
five models and flattening in five, whereas \(D_T(J_0)\) is positive in seven and negative in three.  A local output norm
therefore does not explain the directional differences; this does not show that \(pV\) is causally irrelevant.

\paragraph{Mass-partition restoration.}
Restoring the mass split changes the signed contrast (Figure~\ref{fig:geometry}C; Table~\ref{tab:appx-ggap}).  In
Qwen2.5-1.5B \(D_T(J_0)\) moves from \(+0.00170\) to \(+0.000277\) \([-0.000506,+0.00102]\), with \(G_{\mathrm{gap}}=+0.00142\)
\([+0.000797,+0.00208]\).  At \(J_0\) the seven models with positive original contrasts all have \(G_{\mathrm{gap}}\) intervals
above zero, with very different magnitudes; Gemma-2-2B lies on a separate scale (\(G_{\mathrm{gap}}(J_0)=0.263\)
\([0.245,0.283]\), rescued contrast \(-0.0169\) \([-0.0362,+0.00183]\)).  Llama-3.2-1B stays negative after rescue
(\(-0.00237\) \([-0.00331,-0.00142]\), \(G_{\mathrm{gap}}(J_0)=-0.000758\) \([-0.00157,-0.00000474]\)), and the two larger models'
\(G_{\mathrm{gap}}(J_0)\) intervals cross zero (Qwen2.5-72B \(0.000417\) \([-0.000247,0.00110]\), Llama-3.1-70B \(0.000559\)
\([-0.00102,0.00219]\)).  At \(10J_0\) all ten \(G_{\mathrm{gap}}\) intervals lie above zero (Table~\ref{tab:appx-ggap}), but a
positive signed change is not a smaller absolute gap: Llama-3.2-1B and Qwen2.5-72B move from unresolved contrasts to
resolved negative ones, Qwen2.5-1.5B and Gemma-2-2B also end resolved negative, and Llama-3.2-3B and Llama-3.1-70B
stay resolved positive.  Rescue is thus neither universal nor a unique causal mediation result.

\begin{table}[!htbp]
\centering
\scriptsize
\caption{\textbf{Primary endpoint \(G_{\mathrm{gap}}(10J_0)\).}  Rescued contrast \(D_{\mathrm{rescue}}(10J_0)\) and the signed change \(G_{\mathrm{gap}}(10J_0)=D_T(10J_0)-D_{\mathrm{rescue}}(10J_0)\) (original \(D_T(10J_0)\) in Table~\ref{tab:appx-dt}); 97 blocks, paired block bootstrap of the per-block difference, 5,000 replicates, seed 0.  A positive \(G_{\mathrm{gap}}\) is not necessarily a smaller \(|D|\).}
\label{tab:appx-ggap}
\begin{tabular}{@{}lll@{}}
\toprule
Model & \(D_{\mathrm{rescue}}(10J_0)\) [95\% CI] & \(G_{\mathrm{gap}}(10J_0)\) [95\% CI] \\
\midrule
Qwen2.5-0.5B & 0.00249 [$-$0.000190, 0.00518] & 0.0275 [0.0252, 0.0297] \\
Llama-3.2-1B & $-$0.00862 [$-$0.0112, $-$0.00604] & 0.00842 [0.00657, 0.0103] \\
Qwen2.5-1.5B & $-$0.00278 [$-$0.00510, $-$0.000565] & 0.0161 [0.0144, 0.0178] \\
Gemma-2-2B & $-$0.0512 [$-$0.0854, $-$0.0175] & 0.857 [0.825, 0.889] \\
Qwen2.5-3B & 0.00204 [$-$0.0000977, 0.00417] & 0.0175 [0.0156, 0.0194] \\
Llama-3.2-3B & 0.00371 [0.00142, 0.00607] & 0.00700 [0.00491, 0.00916] \\
Mistral-7B-v0.3 & 0.000505 [$-$0.00106, 0.00199] & 0.0178 [0.0158, 0.0197] \\
Llama-3.1-8B & $-$0.000551 [$-$0.00255, 0.00150] & 0.00787 [0.00593, 0.00973] \\
Llama-3.1-70B & 0.00750 [0.00284, 0.0124] & 0.0206 [0.0158, 0.0256] \\
Qwen2.5-72B & $-$0.00625 [$-$0.00791, $-$0.00471] & 0.00687 [0.00487, 0.00890] \\
\bottomrule
\end{tabular}
\end{table}

\paragraph{Allocation-linked leakage.}
Nine models leak slightly more out of \(S_{0.9}\) under \(R=4\) than under \(R=1\) (differences \(0.000667\) to \(0.001108\);
Qwen2.5-1.5B \(+0.0019\) against \(+0.0012\)), while Llama-3.1-70B leaks slightly less (\(-0.000398\)); all ten
\(D_R(21)\) point estimates still favour \(R=4\) (Figure~\ref{fig:geometry}D).  Because the allocation benefit appears
under both signs of leakage change, it is not explained by retention inside one coarse top-mass set; the leakage
differences carry no intervals, so this is a negative control only.

\subsection{Reconstruction and Rowmax-PoT}
\label{sec:results-secondary}\label{app:rowmax}
\AppendixFigure{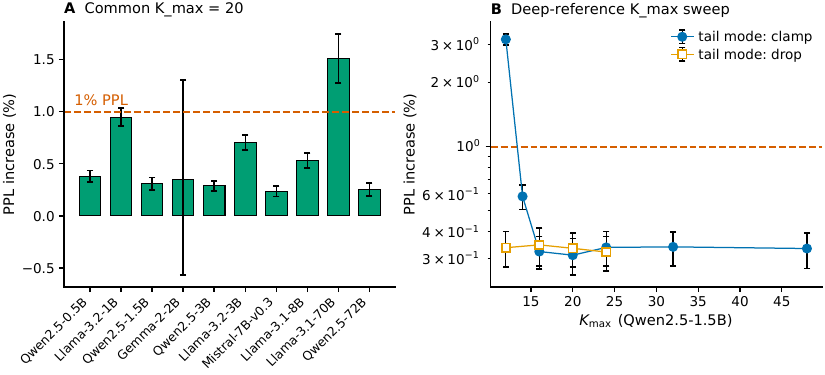}
  {Rowmax-PoT across models and tail cutoffs.}
  {(A)~Ten-model perplexity increase of the Rowmax-PoT \(K_{\max}=20\) clamp condition with 95\% intervals and the 1\% line;
  Llama-3.1-70B is the 1.513\% exception.  (B)~Qwen2.5-1.5B \(K_{\max}\) sweep, clamp and drop tail modes.  Supporting
  characterization, not the Rowmax-H15 kernel result.}
  {fig:rowmax-pot-supp}

At \(K=21\), \(R=4\), upper-edge, nearest-boundary and weight-domain interpolation give \(\Delta\mathrm{NLL}=0.00723\)
(\(0.726\%\) perplexity), \(0.00138\) (\(0.138\%\)) and \(0.0000808\) (\(0.00808\%\)) on the same 97 blocks; the
reconstruction conditions have no pairwise inferential endpoint, so no pairwise intervals are given.  Each rule was
audited against its own local envelope (Appendix~\ref{sec:methods-reconstruction}); these bound the local log-weight error,
not downstream NLL.  Interpolation needs the position within the interval, metadata and compute, and it does not produce
finitely many output values; the ordering holds for this grid and does not establish a general optimum.

Per-model Rowmax-PoT costs at \(K_{\max}=20\) are in Table~\ref{tab:crossmodel-summary} (below 1\% in 9/10 models; Llama-3.1-70B
1.513\%) and, with intervals, in Figure~\ref{fig:rowmax-pot-supp}.  The Qwen2.5-1.5B clamp sweep gives 3.183\%, 0.584\%,
0.322\%, 0.310\%, 0.337\%, 0.339\% and 0.333\% at \(K_{\max}=12\), 14, 16, 20, 24, 32 and 48; dropping instead of clamping the
tail gives 0.336\% at 12 and 0.333\% at 20.  The underlying no-cutoff octave rounding stays within its bound: the
empirical maximum \(|\epsilon|\) matches \(\ln2/(2m)\) for \(m\in\{0.5,1,2,4\}\), the FP32 maxima exceeding the ideal envelope by
\(7.27\times10^{-6}\) to \(2.65\times10^{-5}\) nats, within the recorded tolerance \(\max\{2\times10^{-4}B_m,5\times10^{-5}\}\);
for \(m=1\), 90\% of the attention mass lies within \(\Delta\leq5.5\) nats.  Rowmax-PoT is a numerical example that leads
to the kernel of Appendix~\ref{sec:realization}; it is not evidence of speed-up, energy saving, a complete bit cost or
removal of the \(QK^\top\) product, and it selects no optimal index.

\FloatBarrier
\section{Focused comparison with related work}
\label{app:related}
\label{sec:rw-efficient}\label{sec:rw-probabilities}\label{sec:rw-approx-exp}\label{sec:rw-pot}\label{sec:rw-kernels}
\setcounter{table}{0}\renewcommand{\thetable}{E\arabic{table}}\renewcommand{\theHtable}{E\arabic{table}}

Efficient attention changes \emph{which} score terms are computed \citep{efficient-attention-survey-2025}, and model
compression reduces the precision or size of the representations \citep{transformer-compression-survey-2024}; we keep the
computed terms and ask which properties of the softmax computation a frozen decoder needs.  Table~\ref{tab:rw-compare} places the closest methods.

\begin{table}[!htbp]
\centering
\scriptsize
\setlength{\tabcolsep}{3pt}
\caption{\textbf{Closest methods.}  What each approximates, where its lattice or reference is anchored, what is
calibrated or trained, and where it is evaluated.}
\label{tab:rw-compare}
\begin{tabular}{@{}>{\raggedright\arraybackslash}p{0.13\linewidth}>{\raggedright\arraybackslash}p{0.25\linewidth}>{\raggedright\arraybackslash}p{0.16\linewidth}>{\raggedright\arraybackslash}p{0.18\linewidth}>{\raggedright\arraybackslash}p{0.22\linewidth}@{}}
\toprule
Method & Object & Anchor & Calibration / training & Evaluation \\
\midrule
FQ-ViT & post-softmax probabilities, 4-bit \(\log_2\) & absolute (\(p=1\)) & none & ViT accuracy; no fused kernel \\
RepQ-ViT & post-softmax probabilities, \(\log_{\sqrt2}\) (\(\log_2\) at inference) & absolute, layer-wise scale & percentile calibration & ViT accuracy; no fused kernel \\
ITA & integer streaming softmax, power of two of the floor-rounded distance & row maximum (streaming) & clipping threshold by QAT & 22\,nm ASIC \\
EXAQ & coarse exponential & row maximum & clipping range from activation statistics & frozen LLMs; isolated softmax timing \\
IntAttention & 32-entry exponential lookup table & row maximum & clipping range fixed offline & frozen LLMs; integer attention on Armv8 CPUs \\
FA4 & cubic \texttt{exp2} emulation for part of each row & running maximum (exact exponential) & none; pointwise error & fused kernel on B200 \\
EFQ-Softmax & E2M1 codes \(\{1,1.5\}\times2^e\), affine thresholds & microscaling-block maximum & offline two-parameter search & vector stage of a fused kernel on an A5 unit; zero-shot means \\
Rowmax-H15 (ours) & \(\{1,1.5\}\times2^k\), round-to-nearest-even & attention-row running maximum & none & inside FA4 on B200; NLL, call latency, energy \\
\bottomrule
\end{tabular}
\end{table}

\paragraph{Quantized attention probabilities.}
Post-training quantization of vision transformers found the post-softmax map unusually non-uniform: PTQ4ViT uses a twin
uniform grid \citep{ptq4vit-2022}, FQ-ViT a calibration-free 4-bit \(\log_2\) grid fixed at \(p=1\) \citep{fq-vit-2022}, and
RepQ-ViT a percentile-calibrated \(\log_{\sqrt2}\) grid used as \(\log_2\) at inference \citep{repq-vit-2023}.  Both
logarithmic grids are anchored to a row-independent scale, so their phase relative to the row maximum drifts across rows;
RepQ-ViT noted one consequence (values in 0.354--0.707 share a \(\log_2\) level) and halved the spacing, staying on the
resolution axis.  Our anchor-by-resolution controls (Appendix~\ref{sec:realization-bridge}) add the phase coordinate; they
measure the phase cost of a grid structure in a frozen LLM, not a loss of either method (both evaluated on image
classification).  Row-max-relative quantization of the exponent is itself not new (ITA, EXAQ,
IntAttention), and we make no priority claim.  RepQ-ViT's argument against clipping the largest post-softmax values agrees
with our support results (Appendix~\ref{sec:results-amount}).

\paragraph{Approximate exponentials.}
I-BERT fits a second-order integer polynomial and fine-tunes \citep{i-bert-2021}, I-ViT's Shiftmax trains a base-2
linear approximation \citep{i-vit-2023}, and Softermax uses base 2 with low-precision online normalization in a custom unit
\citep{softermax-2021}.  ITA's integer streaming softmax (22\,nm ASIC) already weights
each score by a power of two of its floor-rounded quantized distance to the streaming row maximum, with the clipping
threshold obtained by quantization-aware training \citep{ita-2023}.  Rowmax-PoT shares the octave lattice but rounds to
nearest without training, and Rowmax-H15 halves the spacing.  EXAQ and IntAttention report small losses in frozen LLMs under
coarse exponentials referenced to the row maximum; their clipping range is calibrated from activation statistics or fixed
offline \citep{exaq-2024,zhong2026intattention}.  We separate which numerical properties that tolerance rests on and
evaluate a specialization with no calibrated range inside a fused GPU kernel.  The relation of Rowmax-H15 to
\citeauthor{schraudolph1999fast}'s \citeyearpar{schraudolph1999fast} bit-level construction, whose offset is chosen
analytically without model data, is given in Appendix~\ref{app:schraudolph}.  Such approximations are commonly justified
by an error budget: I-BERT's polynomial deviates by at most \(1.9\times10^{-3}\), below the 8-bit quantization error of
\(3.9\times10^{-3}\), and FA4 argues the same way for its cubic emulation.  Rowmax-H15 is outside this class---its
per-element \(\log_2\) error spans \([-0.25,0.335]\) octave, far above one BF16 unit in the last place---so we derive its
plausibility from the numerical characterization and validate it at model level.

\paragraph{Power-of-two scales, allocation and operand quantization.}
P\(^2\)-ViT applies power-of-two scaling to requantization scale factors and reuses FQ-ViT's Log-Int-Softmax
\citep{p2-vit-2024}, so it does not place probabilities on a rowmax-anchored lattice; LRP-QViT shows that the allocation of bits
across layers changes accuracy \citep{lrp-qvit-2024}, whereas our allocation is within one score row.  FlashAttention and
FlashAttention-2 establish the fused online-softmax kernel whose running maximum our lattice anchors on
\citep{flashattention-2022,flashattention2-2023}.  FlashAttention-3 quantizes operands to FP8 with block scaling, validated
by pointwise error \citep{flashattention3-2024}; SageAttention 1--3 and INT-FlashAttention quantize the matrix-multiply
operands and keep softmax in full precision
\citep{sageattention-2024,sageattention2-2024,sageattention3-2025,int-flashattention-2024}; MXAttention computes
\(\exp(S-m)\) exactly and quantizes the unnormalized exponentials to MXFP4, data-free \citep{mxattention-2026}.  This family
changes the GEMM operands; we change the nonlinear primitive between them, which is composable in principle but not
evaluated.

\paragraph{FlashAttention-4, the controlled baseline.}
FA4 is the kernel we modify \citep{flashattention4-2026}.  On B200 the exponential unit delivers 16 operations per clock per
SM against 8,192 FP16/BF16 matrix-multiply operations.  FA4 therefore evaluates 10--25\% of each row's exponentials with a
cubic polynomial on the FMA pipeline and reconstructs the result by shifting the integer part into the exponent field, the
same reconstruction step as ours on a different grid.  It aims at accurate \texttt{exp2}, validated pointwise (maximum relative
error \(8.8\times10^{-5}\) in FP32) without a model-level metric, which suits an exact kernel; ours departs from exactness
and therefore reports model-level fidelity.  FA4's conditional rescale (\(\tau=8.0\)) recovers the exact denominator for an
exact exponential but moves the anchor of a lattice; we measure its cost (Appendix~\ref{sec:realization-fidelity}).

\paragraph{Concurrent work.}
EFQ-Softmax \citep{efq-softmax-2026} also removes the dense per-element exponential inside a FlashAttention-style loop, and
the positive part of its E2M1 code set, \(\{1.0,1.5\}\times2^e\), coincides with our half-octave lattice.  E2M1 is required
because its downstream \(PV\) product consumes MXFP4 operands, whereas our \(\{1,1.5\}\) multiplier follows from the
characterization and the FP32 bit construction (Appendix~\ref{app:schraudolph}).  EFQ anchors at the microscaling-block
maximum, uses uniformly spaced affine thresholds with two parameters searched offline per task, and reports a 40.33\%
reduction of the vector stage of a fused kernel and zero-shot means against an MXFP4 baseline.  EFQ and MXAttention both state that numerator and denominator must consume the same approximation;
replacing the primitive upstream of both consumers satisfies this by construction.  VFA pre-computes and freezes the
running maximum \citep{vfa-2026}, a schedule change rather than a probability quantizer.  Operand quantization and
MXAttention keep exact exponentials, FA4 approximates \texttt{exp2} to storage precision, and EXAQ and EFQ (calibrated
parameters), IntAttention (range fixed offline) and our work (no calibrated range) coarsen it substantially.

\paragraph{Engineering reports on FA4.}
Low-Precision FA4 \citep{lp-fa4-mxfp8-2026} adds block-scaled MXFP8 operands in training on GB300, leaving softmax and
\texttt{exp2} untouched, and states the mechanism our dtype~\(\times\) mask controls probe: as matrix-multiply throughput
rises, special-function-bound softmax work becomes exposed.  GDPA \citep{gdpa-2026} replaces softmax with element-wise
activations on B200 and an SFU \(\tanh\) with an ALU-only polynomial.  Neither Meta report is peer reviewed or gives a
model-level accuracy number.

\end{document}